\documentclass[11pt]{book}
\usepackage{iiit_thesis}
\usepackage{times}
\usepackage{epsfig}
\usepackage{amsmath}
\usepackage{amssymb}
\usepackage{latexsym}
\usepackage{graphicx}
\usepackage{multirow}
\usepackage{algorithm2e}
\usepackage{amsfonts}
\graphicspath{{images/}{./}} 
\usepackage{soul}
\usepackage{epstopdf}
\usepackage[noadjust]{cite}
\usepackage{nomencl}

\newtheorem{theorem}{\textbf{Theorem}}
\newtheorem{remark}{\textbf{Remark}} 
\newtheorem{prop}{\textbf{Property}} 
\newtheorem{lemma}{\textbf{Lemma}}

\newtheorem{assum}{\textbf{Assumption}}
\counterwithin{mydef}{chapter}
\counterwithin{remark}{chapter}
\counterwithin{problem}{chapter}
\counterwithin{motivation}{chapter}
\counterwithin{assum}{chapter}
\counterwithin{lemma}{chapter}
\counterwithin{theorem}{chapter}
\counterwithin{proposition}{chapter}

\usepackage[resetlabels,labeled]{multibib}
\newcites{related}{Related Publications}

\long\def\symbolfootnote[#1]#2{\begingroup%
\def\thefootnote{\fnsymbol{footnote}}\footnote[#1]{#2}\endgroup}
\renewcommand{\baselinestretch}{1.2}
\onecolumn

\begin{document}
\pagenumbering{roman}

\thispagestyle{empty}
\begin{center}
\vspace*{1.5cm}
{\Large \bf Adaptive Steering Control for Steer-by-Wire Systems}

\vspace*{3.75cm}
{\large Thesis submitted in partial fulfillment\\}
{\large  of the requirements for the degree of \\}

\vspace*{1cm}
{\it {\large (Master of Science in \textbf{Computer Science} by Research} \\}

\vspace*{1cm}
{\large by}

\vspace*{5mm}
{\large Harsh Shukla\\}
{\large 2019701012\\
{\small \tt harsh.shukla@research.iiit.ac.in}}

\vspace*{4.0cm}
{\psfig{figure=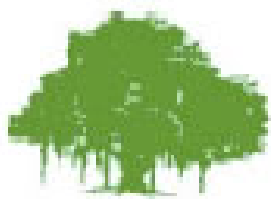,width=14mm}\\}
{\large International Institute of Information Technology\\}
{\large Hyderabad - 500 032, INDIA\\}
{\large August 2021\\}
\end{center}

\newpage
\thispagestyle{empty}
\renewcommand{\thesisdedication}{{\large Copyright \copyright~~Harsh Shukla, 2021\\}{\large All Rights Reserved\\}}
\thesisdedicationpage

\newpage
\thispagestyle{empty}
\vspace*{1.5cm}
\begin{center}
{\Large International Institute of Information Technology\\}
{\Large Hyderabad, India\\}
\vspace*{3cm}
{\Large \bf CERTIFICATE\\}
\vspace*{1cm}
\noindent
\end{center}
It is certified that the work contained in this thesis, titled  by Harsh Shukla, has been carried out under
my supervision and is not submitted elsewhere for a degree.

\vspace*{3cm}
\begin{tabular}{cc}
\underline{\makebox[1in]{15.08.2021}} & \hspace*{5cm}
\underline{\makebox[2.5in]{}} \\

Date & \hspace*{5cm} Adviser: Prof. Spandan Roy
\end{tabular}
\oneandhalfspace

\newpage
\thispagestyle{empty}
\renewcommand{\thesisdedication}{\large Dedicated to SCIENCE, POETRY and MUSIC}
\thesisdedicationpage

\mastersthesis
\renewcommand{\baselinestretch}{1.5}

\chapter*{Acknowledgments}
\label{ch:ack}
I have been influenced my many people in the time it took to complete the work in these pages. First and foremost has been my advisor and mentor, Prof. Spandan Roy. I was fortunate enough to find someone who shares the same passion for self driving cars and how will they will shape our world in future. He helped me in pursuing the problems in this domain and guided me througout the time. Many of the ideas in this thesis were developed at his suggestion. I hope I have done them justice. I have had the pleasure of working in the Robotics Research Center with some great people. I met here some of the brightest minds seen. All these people inspired me and still inspire to do more and reach out for great things. They made the lab a fun and enjoyable place to talk about research, or anything at all. Special thanks to Mithun Nallana and Saransh Dave for their valuable inputs.  Finally, I want to thank my parents and my brother for their patient support throughout all these years. I would like to thank people known me and have always been with me from my undergrad days. I would like to thank Anjali and Dheeraj, for always believing in me. 

\chapter*{Abstract}
\label{ch:abstract}

Steer-by-Wire (SBW) systems are being adapted widely in semi-autonomous and fully autonomous vehicles. The main control challenge in a SBW system is to follow the steering commands in the face of parametric uncertainties, external disturbances and actuator delay;crucially, perturbations in inertial parameters and damping forces give rise to state-dependent uncertainties, which cannot be bounded a priori by a constant. However, the state-of-the-art control methods of SBW system rely on a priori bounded uncertainties, and thus, become inapplicable when state-dependent dynamics become unknown. In addition, ensuring tracking accuracy under actuator delay is always a challenging task.

This work proposes two control frameworks to overcome these challenges. Firstly, an adaptive controller is proposed to tackle the state-dependent uncertainties and external disturbances in a typical SBW system without any a priori knowledge of their structures and of their bounds. The stability of the closed-loop system is studied analytically via uniformly ultimately bounded notion and the effectiveness of the proposed solution is verified via simulations against the state-of-the-art solution. While this proposed controller handles the uncertainties and external perturbations, it does not consider the actuator delay which sometimes result in decreased accuracy. Therefore, a new adaptive-robust control framework is devised to tackle the same control problem of an SBW system under the influence of time-varying input delay. In comparison to the existing strategies, the proposed framework removes the conservative assumption of a priori bounded uncertainty and, in addition, the Razumikhin theorem based stability analysis allows the proposed scheme to deal with arbitrary variation in input delay. The effectiveness of the both controllers is proved using comparative simulation studies. 

\tableofcontents
\listoffigures
\listoftables

\makenomenclature
\renewcommand{\nomname}{List of Symbols}
\renewcommand{\nompreamble}{The list describes several symbols that will be later used throughout the thesis}
\mbox{}
\nomenclature{$\mathbb{R}$}{real Line}
\nomenclature{$\mathbb{R}^{+}$}{real line of Positive Numbers}
\nomenclature{$\mathbb{R}^{n}$}{real space of dimension $n$}
\nomenclature{$\forall$}{for all }
\nomenclature{$\exists$}{there exists}
\nomenclature{$\vert \bullet \vert$}{absolute value of the argument vector/matrix}
\nomenclature{$\mathbb{R}^{n \times n}$}{real space of dimension $n \times n$}
\nomenclature{$\lVert\,{\bullet}\,\rVert$}{Euclidean norm of the argument vector/matrix}
\nomenclature{$\lambda_{\min}(\bullet)$}{minimum eigenvalue of the argument matrix}
\nomenclature{$\lambda_{\max}(\bullet)$}{maximum eigenvalue of the argument matrix}
\nomenclature{$\lor$}{logical OR operation}
\nomenclature{$\land$}{logical AND operation}
\nomenclature{$\Xi$}{positive semi-definite Matrix}
\nomenclature{sgn($\cdot$)}{standard signum Function}
\nomenclature{$I$}{identity matrix of proper dimensions}
\printnomenclature


\chapter{Introduction}
\label{ch:intro}
The word \textit{robot} was originally used as a fictional humanoid character in a 1920’s Czech play. However, the modern meaning of the word `robot’ encompasses not just humanoids but any machine capable of completing a set of tasks autonomously. Often this automation is achieved through sensors, computers, and actuators on the machine. 

Robots are comprised of the systems which can be presented in form of differential equations. Although this mathematical modelling is helpful in simplification of a real world model, often these models are imperfect. These imperfections rise from our biases and assumptions while studying the real world systems. In this thesis, we study these issues in SBW systems and design controllers to mitigate the same. 

\section{Automotive Steering Systems}
Over the past two decades, a growing thrust area of research has been electrification of different mechanical and hydraulic systems in ground vehicles, which offers less environmental concerns due to the removal of hydraulic fluids and continual engine parasitic losses. A steering system  works on the driver's input which is transmitted to the rack and pinion arrangement connected to the wheels through some mechanical or electronic mechanisms. Although basic design of the steering systems have remained same, improvements were in place of the mechanical shafts between the steering and rack. Mechanical shaft was replaced by a hydraulic pump in 1950s, however with the onset of the last decade of 20th century, electronic modifications started making their way in steering systems and power assist was introduced \cite{power-steering}. Since then, power steering systems have become a norm for automobiles. \textit{X-by-Wire} systems have been the most recent development in the electrification of the automobile systems. This term is usually discussed in the context of replacing X with vehicle controls such as braking, acceleration and steering \cite{SBW}. In these systems, the control of the vehicle is supported by electronically assisted controller. These systems have been very useful in self-driving car research.

\section{X-by-Wire} 
\label{sec:X-by-Wire}
A lot of automobile vehicles now are available with X-by-Wire technology along with a mechanical backup. A fully developed X-by-Wire system in future is supposed to replace all the mechanical backups from the vehicles. There are several advantages of X-by-Wire systems over purely mechanical systems: these electronic systems eliminate the need for hydraulic systems and thereby increase the packaging density of the vehicle. 
The electrical equivalents for traditional mechanical linkages and hydraulic power assist systems include ‘brake- by-wire’, ‘steer-by-wire’ and ‘throttle-by-wire’ system. These equivalents, collectively known as ‘X-by-Wire’ technologies, form the fundamental structure for the Semi and Fully Autonomous Driving Systems. 
As compared to a traditional steering system, a Steer-by-Wire (SBW) system removes the mechanical shaft that connects the steering column and steering pinion: therefore, in absence of any physical connection, the steering input is transmitted via a by-wire electronic communication module. 

In order to have the SBW equipped vehicle have the similar features to that of a conventional steering system, it is important to devise a tracking control mechanism for the steering actuator such that the steering commands from hand-wheel are followed as precisely as possible; 
however, in presence of inevitable modelling imprecision and parametric uncertainties, advanced control strategies are eventually investigated. In the following section, we discuss the SBW dynamics, various control problems, state-of-the-art solutions and their issues.

\section{Steer-By-Wire System - Architecture and Dynamics}

\begin{figure}[!h]
	\centering
	\caption{Architecture of a typical steer-by-wire system.}
	\includegraphics[width=3.0in,height=4.0in]{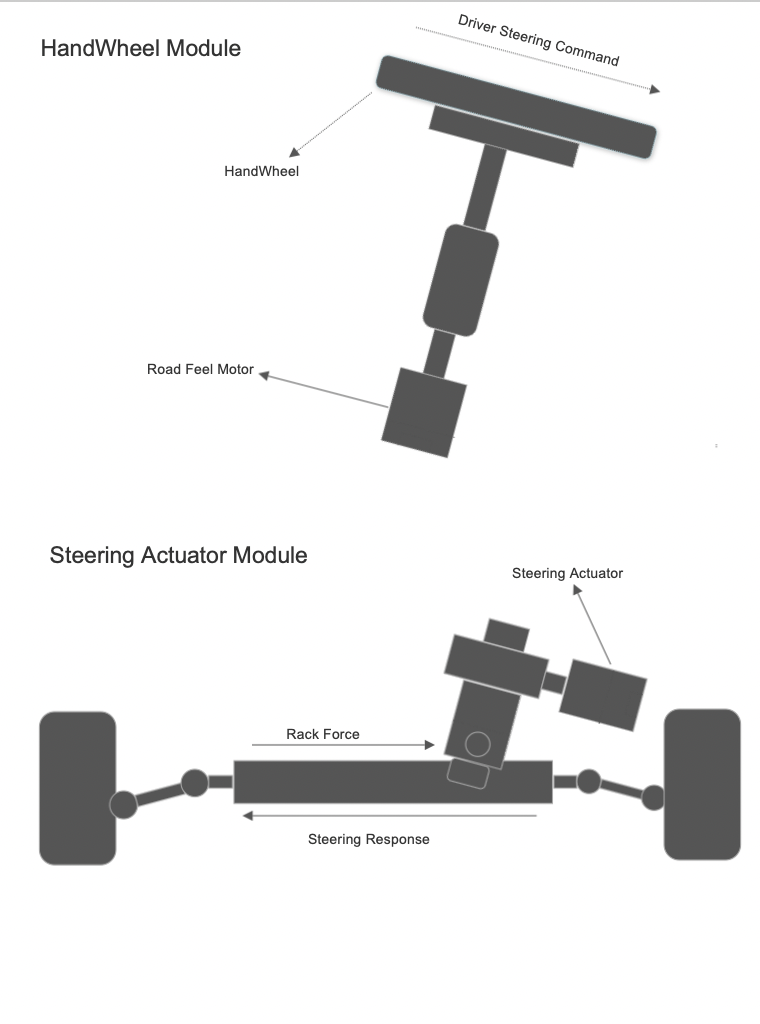}
	\label{fig:sbw}
\end{figure}

\subsection{Architecture}
A typical SBW system is formed by removing the mechanical linkage between the steering column and steering actuator in a conventional power steering
system (cf. Fig. \ref{fig:sbw}). Thereby, a SBW system is made up of two main components, namely, the steering section and the wheel section: (i) the steering section consists of steering wheel, feedback actuator and feedback actuator angle sensor; (ii) the wheel section contains the wheel, rack and pinion, steering actuator and pinion angle sensor. Steering actuator with the input from the steering hand-wheel and feedback from the pinion angle sensor produce the control signal.
To have an accurate steering performance, the key control challenge is to make the front wheel angle follow the hand-wheel reference angle in the presence of disturbances stemming from parametric uncertainties and external disturbances. 

\subsection{Dynamics}
\begin{figure}[!h]
	\centering
	\includegraphics[width=3.0in,height=2.5in]{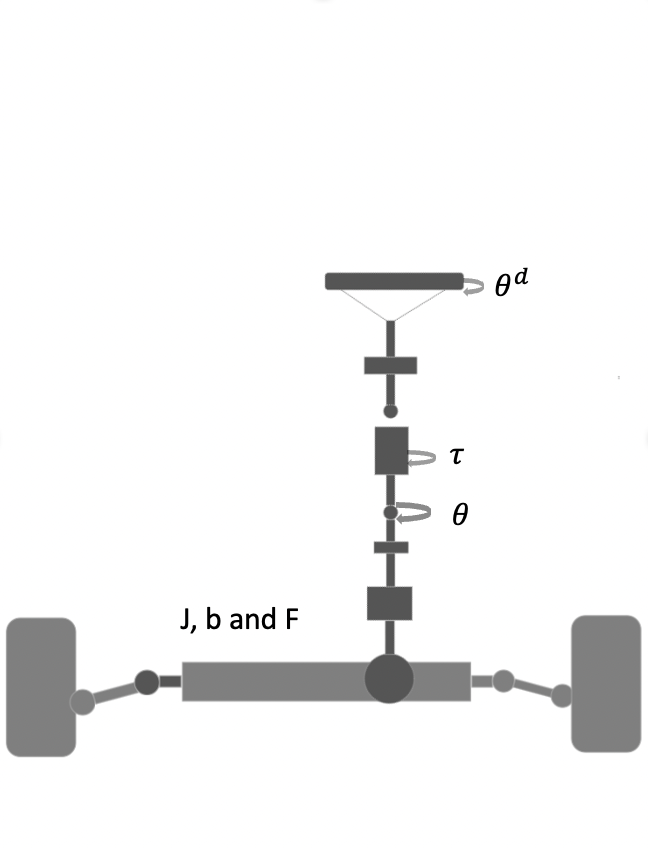}
	\caption{Schematic of a steer-by-wire system for dynamics formulation.}\label{fig:dyn} 
\end{figure}

The dynamics of SBW system can be found via the celebrated Euler-Lagrange (EL) formulation and it has the following form \cite{SBW}: 
\begin{equation}
    J\ddot{\theta} + B\dot{\theta} + F + i_{rc}F_{rack} + \tau_a = \tau, \label{dyn}
\end{equation}
\begin{align*}
\text{where}~~   J &= J_{c}  + i_{gc}^{2}J_{gear} + i_{gc}^{2}J_{m} + i_{rc}^{2}M_{rack} \\
    B &= B_{c} +i_{gc}^{2}B_{gear} + i_{gc}^{2}B_{m} + i_{rc}^{2}B_{rack} \\
  \tau & =    i_{mc}\tau_{m}.
\end{align*}
Here $\theta$ is the angle of steering column; $J$ is the system equivalent inertia; $B$ is the system equivalent viscous damping; $F$ and $F_{rack}$ represent system friction and rack force on steering rod respectively; $\tau_a$ is the influence of tire forces
on the steering system and $\tau$ is the control input. A schematic denoting such dynamics is depicted in Fig. \ref{fig:dyn}, while other system parameters are described in Table \ref{table 1}.

\begin{table}[!t]
\renewcommand{\arraystretch}{1.2}
\caption{System Parameters}
\label{table 1}
\centering
{
\begin{tabular}{ c c }
		\hline
		\hline
		Parameters & Description \\
	 \hline \\
	 $J_{c}$ &  inertia of steering column\\
	 $J_{m}$ &  inertia of  motor\\
	 $J_{gear}$ &  inertia of reduction gearbox \\
	$M_{rack}$ &  mass of steering rack\\
	$i_{rc}$ & generalized transmission ratio
from tie-rod to the steering column shaft \\
	$i_{gc}$ & reduction ratio from gearbox to steering column shaft\\
	$B_{c}$ & viscous damping of the steering column  \\
	$B_{m}$ & viscous damping of the motor\\
	$B_{gear}$ & viscous damping of the reduction gearbox\\
	$B_{rack}$ &  viscous damping of the steering rack\\
	$\tau_m$ & output torque of steering actuator motor\\
		\hline
		\hline
\end{tabular}}
\end{table}


\section{Related Works}

In this chapter, we first review the existing approaches for control design for steering control in uncertain environments in the backdrop of autonomous driving.  Subsequently, we review the more niche approaches specifically designed for such systems targeting their applications in the autonomous driving and other drive-by-wire systems. 

\subsection{Controller Design for SBW Systems}

The SBW systems are inherently uncertain systems. These are second order systems whose dynamics vary with changing road conditions and mechanical parameters. Such systems require the controllers which take into all these perturbations and uncertainties. 
With the aim of making the SBW systems as close to traditional steering systems, the operation from the hand-wheel to the front wheel is considered as a tele-operation application. 
Initial attempts were made to solve such control problem employing conventional PID controller \cite{PID1}. However, these approaches tend to come with the a large inherent overshoot and response delay due to the integral term involved in the control law. Another paradigm which researchers used was designing the controller on the basis of model information.  In \cite{Paul}, a controller was introduced which used acceleration, velocity and feed-forward compensation to improve upon the performance of the SBW controller response. 


For steering control of the SBW systems, the main challenge is to handle the parametric uncertainties and external disturbances, which were not considered in the aforementioned works. The various sources of uncertainties include unmodelled friction forces, external rack force and other uncertainties in the model parameters. To tackle uncertainties in SBW systems, researchers have primarily applied robust control \cite{wang2014robust, wang2013sliding} and adaptive control methods \cite{ATC, 5272130, kim2020adaptive, cetin2012adaptive, zhang2020adaptive, sun2015robust}. While the adaptive control solutions do not require a priori knowledge of the bound of uncertainties compared to a robust controller, a few observations worth mentioning: the state-of-the-art adaptive control methods either require structural knowledge of uncertainty \cite{cetin2012adaptive, sun2015robust}, or assume the uncertain dynamics to be bounded by a constant a priori \cite{ATC, kim2020adaptive, zhang2020adaptive}. However, the upper bound of uncertain dynamics of a SBW system has explicit dependency on states \cite{cetin2012adaptive}; under such an uncertainty setting, presumption of constant upper bound on uncertainty is not only conservative \cite{roy2019overcoming, roy2019simultaneous, roy2019, rodrigues2018global, roy2020towards, roy2018new, sankaranarayanan2020aerial, tao2020issue, roy2021adaptive}, it may even lead to instability \cite{cst_new, roy2020adaptive, tao2020stable}.   


\subsection{Controller Design under Time Delay}
Time delays can be encountered in real-life systems either naturally in the form of communication delay, actuator delay \cite{Ref:1, Ref:3, roya2020wide} or artificially via a part of control design \cite{roy2018analysis, roy2015robust, roy2014time, roy2013robust, galip2015time, ramirez2015design,roy2021artificial,roy2020time}. Left uncompensated, time delays may potentially compromise controller accuracy and system stability. Therefore, the study of stability and performance of dynamical systems under the influence of time delay have recieved extensive attention for many years. The state-of-the-art in this domain is well documented in \cite{Ref:1, Ref:2, Ref:3, Ref:4,Ref:5} along with the existing challenges. In SBW systems, actuator delay can cause the controllers to perform poorly which might reflect on the performance of the vehicle. Amongst the earliest controllers, a few notable works include Smith predictor based approach \cite{Ref:6}, Artstein model reduction \cite{Ref:7}, finite spectrum assignment \cite{Ref:8}, predictor based methodology \cite{Ref:8, Ref:9, Ref:10, Ref:11, Ref:12, Ref:13, Ref:19, Ref:27}. 
The case of feedforward linear systems under both state and input delay is reported by \cite{Ref:14}. Further, to tackle system uncertainties various robust and adaptive control solutions are proposed in \cite{Ref:16, Ref:17, Ref:29, Ref:30}. Unfortunately, the adaptation mechanism in \cite{Ref:30} yields a monotonically increasing switching gain, which might cause high gain instability \cite{plestan2010new, shtessel2012novel,roy2020vanishing, roy2019role, ioannou2012robust}. 

In comparison, the literature for nonlinear systems with input delay is relatively not vast and thus, more challenging. The earlier works, \cite{Ref:20,Ref:21, Ref:24,Ref:41} do not consider parametric uncertainties. This issue was addressed in \cite{Ref:31} via a predictor based approach for constant input delay, which was later extended for time varying input delay in \cite{Ref:32} and also for state delay \cite{Ref:33}. However, these require exact knowledge of input delay and can only negotiate slowly varying input delay. A robust controller is proposed in \cite{roy2017robust} to tackle time varying input delay via Razumikhin theorem based analysis. However, a priori knowledge of uncertainty bounds is required to design any robust controller, which is not always available in practice due to unmodelled dynamics and unknown disturbances. 

To handle the situation of unknwon uncertainty bound, adaptive-robust control laws are developed (cf. \cite{utkin2013adaptive, roy2017adaptive, Ref:self, roy2020adaptive, roy2018adaptive, roy2017adaptive2, roy2017adaptive4} and references therein). To compensate time varying network delays in uncertain complex network systems, adaptive sliding mode based control law is developed in \cite{Ref:44}. In a different application, \cite{Ref:45} used adaptive law to evaluate the coupling strength for synchronization of chaotic systems under network induced delay. However, the adaptive laws propsed in \cite{Ref:44, Ref:45} create overestimation of switching gains by not allowing these gains to decrease \cite{roy2019overcoming}. The issue of over- and under-estimation problems of switching gain for uncertain Euler-Lagrange systems under arbitrarily varying input delay was first addressed in \cite{Ref:self}. On the other hand, this work considers the uncertainties to be bounded by an unknown constant, which is a conservative assumption \cite{utkin2013adaptive} when system states are involved in the upper bound structure of uncertainty (cf. \cite{roy2019overcoming, roy2019simultaneous, roy2018new, roy2020adaptive}). 

\section{Contributions}
SBW systems are considered an important aspect in the development of self-driving vehicles. It becomes essential to design controllers for such systems which can handle the uncertainities stemming from system parameters, imprecise modelling and actuator delay. This thesis contributes in this pursuit in the following manner: 
\begin{itemize}
    \item A new adaptive controller for SBW systems is proposed to tackle state-dependent uncertainties that are not a priori bounded. The proposed design does not require any knowledge of structure and bound of uncertainties. 
    \item To address the nonlinearities arising from actuator delay, an adaptive control framework is proposed based on the Razumikhin-theorem which can address input delay of unknown variation along with state-dependent uncertainties. 
\end{itemize}

We validate both of these controllers via simulations with different velocity and road conditions to expose the system with various uncertainties.

\section{Organization of Thesis}
This thesis is organized into four chapters. The summary of the work presented in each chapter is briefly outlined as follows: 

\textbf{Chapter 1}: This introductory chapter elaborates the motivation of this research, problem orientation, the pertaining gaps in literature, the main contributions and an outline of the thesis.

\textbf{Chapter 2}: This chapter proposes adaptive control framework that can tackle the state-dependent uncertainties and external disturbances in a typical SBW system without any a priori knowledge of their structures and of their bounds. The stability of the closed-loop system is studied analytically via uniformly ultimately bounded notion and the effectiveness of the proposed solution is verified via simulations against the state-of-the-art solution. 

\textbf{Chapter 3}: This chapter introduces a new adaptive control strategy to address unknown time varying actuator delay along with state-dependent uncertainties. The closed-loop stability was analysed via Lyapunov-Razumikhin theorem to find maximum allowable time delay. The efficacy of the proposed methodology is substantiated in the simulated environments. 

\textbf{Chapter 4}: This chapter presents the conclusion of the thesis with a concise summary of primary contributions and a brief discussion about future research work.


\chapter{Adaptive Control of Steer-by-Wire Systems under Unknown State-dependent Uncertainties}
\label{ch:chap3}

The assumption that uncertainty is upper bounded by a constant does not hold true when the uncertainty has explicit dependency on the states of the system. This restricts the state to be an upper bounded prori, and due to this the value of stability analysis is undermined. In the light of the above assumption and the discussions in the previous chapter (cf. Sect. 1.4.1), an adaptive control solution that can tackle unknown dynamics of SBW system is still missing. Toward this direction, the proposed adaptive control mechanism has the following major contributions:
\begin{itemize}
	\item The proposed control framework avoids any a priori bounded assumption on the state-dependent uncertain dynamics.
	
	\item Compared to the state-of-the-art (cf.  \cite{ATC, 5272130, kim2020adaptive, cetin2012adaptive, zhang2020adaptive, sun2015robust}), the proposed adaptive law does not require any a priori knowledge of the structure of uncertain system dynamics and of parameters. 
	
	\end{itemize}
The closed-loop system stability is analysed using Uniformly Ultimately Bounded notion and the effectiveness of the proposed design is validated via comparative simulation studies. 

The remainder of this chapter is organized as follows: the system dynamics and control problem  is formulated in Section 2.1; Section 2.2 details the proposed controller design. Section 2.3 presents the closed-loop stability analysis for the proposed controller. Finally, Section 2.4 presents the simulation results.


\section{System Dynamics and Assumptions}
The dynamics of SBW system can be found via the celebrated Euler-Lagrange (EL) formulation and it has the following form \cite{SBW}: 
\begin{equation}
    J\ddot{\theta} + B\dot{\theta} + F + i_{rc}F_{rack} + \tau_a = \tau, \label{dyn}
\end{equation}
\begin{align*}
\text{where}~~   J &= J_{c}  + i_{gc}^{2}J_{gear} + i_{gc}^{2}J_{m} + i_{rc}^{2}M_{rack} \\
    B &= B_{c} +i_{gc}^{2}B_{gear} + i_{gc}^{2}B_{m} + i_{rc}^{2}B_{rack} \\
  \tau & =    i_{mc}\tau_{m}.
\end{align*}
Here $\theta$ is the angle of steering column; $J$ is the system equivalent inertia; $B$ is the system equivalent viscous damping; $F$ and $F_{rack}$ represent system friction and rack force on steering rod respectively; $\tau_a$ is the influence of tire forces
on the steering system and $\tau$ is the control input. A schematic denoting such dynamics is depicted in Fig. \ref{fig:dyn}, while other system parameters are described in Table \ref{table 1}.

In the following, we highlight some system properties and the amount of uncertainty present in the system dynamics:
\begin{prop}
There exist positive scalars $f_1, f_2, f_3$ such that $|F | \leq f_1 $, $| F_{rack} | \leq f_2, |\tau_a | \leq f_2$.
\end{prop}
\begin{remark}
It is worth mentioning here that, we have avoided state-dependent upper bound structure of the friction term $F$ as the viscous friction/damping part is considered separately. Therefore, we do not deviate from the general property for damping forces of an EL system \cite{spong2008robot}. Whereas, the upper bound structures for $ F_{rack}$ and $\tau_a$ are standard for a SBW system \cite{SBW}. 
\end{remark}
\begin{assum}[Uncertainty]
The system dynamics parameters $J, B$ and the upper bounds $f_1,f_2,f_3$ are unknown.
\end{assum}
Property 1 and Assumption 2.1 act as a control design challenge since it deviates from assumptions often found in the state-of-the-art control designs: a few observations in this direction are as follows:
\begin{remark}[On the state-dependent uncertainty]\label{rem_state}
While dealing with parametric uncertainties, it is often assumed that the lumped uncertainties are bounded a priori by a constant (cf. \cite{ATC, kim2020adaptive, zhang2020adaptive} and references therein). However, it can be noted from (\ref{dyn}) that perturbations in $J$ and $B$, say $\Delta J$ and $\Delta B$ respectively, create a state-dependent lumped uncertainty $(\Delta J \ddot{\theta} + \Delta B \dot{\theta})$, which cannot be bounded a priori; it is shown in \cite{cst_new, roy2020adaptive} that, adaptive control designs based on such a consideration may lead to unstable behaviour. In this work, we avoid such assumption.
\end{remark}
\begin{remark}[On a priori knowledge of system]\label{rem_2}
In contrast to the existing methods, no a priori knowledge of system parameters $J$ and $B$ (cf. \cite{SBW}) and of structure (or model) of friction $F$ (cf. \cite{cetin2012adaptive, sun2015robust}) is required. 
\end{remark}
In a SBW system, the objective is to make the angle steering column $\theta(t)$ follow some predefined desired trajectory $\theta^d(t)$ as close as possible. In this direction, the following standard assumption is taken:
\begin{assum}
The desired angle of steering column $\theta^{d}(t)$ and its time derivatives $\dot{\theta}^{d}(t),\ddot{\theta}^{d} (t)$ are assumed to bounded, i.e., $\theta^{d}, \dot{\theta}^{d},\ddot{\theta}^{d} \in L_{\infty}$.
\end{assum}

In view of the above discussions, the control problem can be defined as follows: 

\textbf{Control Problem:} Under Assumptions 2.1-2.2 and Property 1, to design an adaptive control solution for SBW system (\ref{dyn}), perturbed by state-dependent uncertainties and without any a priori knowledge of system parameters in line with Remarks \ref{rem_state} and \ref{rem_2}.

In the following section, we solve the control problem. 
\section{Controller Design}

Let us define the trajectory tracking error as $e(t) = \theta(t) - \theta^{d}(t)$. We further define a filtered tracking error variable $r$ as
\begin{equation}
 r (t)\triangleq \dot{ e}(t)+ \lambda e(t) , \label{r} 
\end{equation}
where $\lambda \in \mathbb{R}^{+}$ is a user-defined design parameter. In the following, we shall omit variable dependency for compactness. 

Multiplying the time derivative of (\ref{r}) by $ J$ and using (\ref{dyn}) yields
\begin{align}
 {J}\dot{ r}&=  J(\ddot{ \theta}-\ddot{ \theta}^d+ \lambda\dot{ e})\nonumber \\
 &= \tau + \varphi, \label{r dot}
\end{align}
where 

\begin{align}
    \varphi \triangleq -( B \dot{ \theta}+ F + i_{rc}F_{rack} + \tau_a +  {\ddot{ \theta}^d} -  J \lambda\dot{ e}) \label{un}
\end{align}
represents the overall uncertainty. To design the proposed controller, it is important to derive the upper bound structure of $| \varphi |$, and it is carried out subsequently.

\textbf{Upper Bound Structure of $| \varphi |$:} Using Property $1$, from (\ref{un}) we have
\begin{align}
| \varphi |& \leq B | \dot{ \theta} | + f_1 +i_{rc} f_2 +f_3 + J (| \ddot{ \theta}^d |+  \lambda  |\dot{ e} | ). \label{new}
\end{align}

\noindent Further, let us define $\xi \triangleq [ e ~ \dot{ e}]^T $. Then, substituting $\dot{{\theta}} = \dot{{e}} + \dot{ \theta}^d$ into (\ref{new}), using inequality $|| \xi|| \geq | \dot{ e} | $ and Assumption 2.2 yields 
\begin{align}
| \varphi | &\leq K_{0 }^{*} + K_{1 }^{*}  ||   \xi || , \label{psi_bound} \end{align}
\begin{align*}
\text{where}~~ K_{0 }^{*} \triangleq &  B | \dot{ \theta}^d | + f_1 +i_{rc} f_2 +f_3 + J | \ddot{ \theta}^d |, \\
 K_{1 }^{*} \triangleq & B+J  \lambda
\end{align*}
are \textit{unknown} finite scalars. 

Hence, a state-dependent (via $\xi$) upper bound structure in uncertainty can be observed for SBW system from (\ref{psi_bound}) and based on this, we propose a control law as

\begin{align}
&\tau = - \gamma  r - e - \rho sat({r}), ~~sat(r) = \begin{cases}
r/|r| & ~\text{if}~|r| \geq \epsilon\\
r/\epsilon &~\text{if}~ |r| < \epsilon
\end{cases}
\label{input rob}\\
&  {\rho}=\hat{K}_{0} + \hat{K}_{1} ||  \boldsymbol{ \xi} ||  , \label{rho} 
\end{align}
where $\gamma $ is a positive definite user-defined scalar and $\epsilon \in \mathbb{R}^{+}$ is used to avoid discontinuity in the control law. 
The gains $\hat{K}_{i}$ are evaluated via the adaptive laws: 
\begin{subequations}\label{split_adap}
\begin{align}
& \dot{\hat{K}}_{0} =| {r} |  - \alpha_{0}\hat{K}_{0}, \label{hat_k_0} \\
&\dot{\hat{K}}_{1} =| {r}| || \xi || - \alpha_{1}\hat{K}_{1}, \label{hat_k_1}\\
\quad \text{with}~~& \hat{K}_{0}(0)>0,~\hat{K}_{1} (0)>0 \label{init} 
\end{align}
\end{subequations}
where $\alpha_{i} \in \mathbb{R}^{+}$, $i=0,1$ are design scalars. A few observations are as follows regarding the proposed solution over the state of the art.

\begin{remark}
Owing to the structures of the error variable $r$ as in (\ref{r}) and of saturation based robustness component in control input as in (\ref{input rob}), the proposed design can also be considered as a kind of adaptive sliding mode control. Nevertheless, compared to the state-of-the-art solutions, the proposed method (i) does not require any knowledge of system dynamics parameters (cf. \cite{roy2019overcoming, plestan2010new, shtessel2012novel}); (ii) does not presume a priori bounded uncertainty (cf. \cite{edwards2016adaptive, utkin2013adaptive, oliveira2016adaptive, moreno2016adaptive, zhang2018adaptive, plestan2010new, shtessel2012novel}) and (iii) does not use increasing-decreasing type non-smooth adaptive law (cf. \cite{roy2019overcoming, roy2018adaptive}).  
\end{remark}

\section{Stability Analysis of The Proposed Controller}
\begin{theorem}
	Under Property 1 and Assumption 2.1-2.2, the closed-loop trajectories of (\ref{r dot}) using the control laws (\ref{input rob}) and (\ref{rho}) in conjunction with the adaptive law (\ref{split_adap}) are Globally Uniformly Ultimately Bounded (GUUB). Further, an ultimate bound $\omega$ on the tracking error variable $\xi$ is found to be 
\begin{align}
\omega = \sqrt{ \frac{\varsigma+\sum_{i=0}^{1} {\alpha_i{K_{i}^{*}}^2}}{ \left( \varrho - \kappa \right)}} , \label{ub_4}
\end{align}
where $\varrho \triangleq \frac{\min_{ i} \lbrace \gamma, ~\lambda,~ \alpha_i/2 \rbrace} {\max \lbrace J/2,~ 1/2  \rbrace}$, $0< \kappa < \varrho$ and $\varsigma>0$ is a constant defined during the proof (cf. \ref{var}).
\end{theorem}
\textit{Proof.}
The solutions of adaptive laws (\ref{hat_k_0})-(\ref{hat_k_1}) with initial condition in (\ref{init}) can be obtained as 
\begin{align}
\hat{K}_i(t) &= \underbrace{\exp(-\alpha_i t) \hat{K}_i (0)}_{\geq 0} + \underbrace{\int_{0}^{t} \exp(-\alpha_i(t-\psi)) (|| {r} (\psi)|| || \xi (\psi)||^{i}) \mathrm{d}\psi}_{\geq 0},~~ i=0,1
\end{align}
leading to
\begin{align}
 &\hat{K}_{0} (t)  \geq 0, ~ \hat{K}_{1} (t)  \geq 0 ~~\forall t\geq 0. \label{low_bound}
\end{align}
Closed-loop stability is analyzed using the Lyapunov function:
\begin{align}
V& = \frac{1}{2}  J{r^2} + \frac{1}{2}e^2 + \sum_{i=0}^{1}  \frac{1}{2}(\hat{K}_{i} -{K}_{i}^{*})^2 . \label{lyap}
\end{align}
Using (\ref{r}), (\ref{r dot}) and (\ref{input rob}), the time derivative of (\ref{lyap}) yields 
\begin{align}
\dot{V}&= r J\dot{ r} + e\dot{e} + \sum_{i=0}^{1}(\hat{K}_{i} -{K}_{i}^{*})\dot{\hat{K}}_{i} \nonumber \\
& = r(\tau + \varphi)+ e (r- \lambda e) +\sum_{i=0}^{1}(\hat{K}_{i} -{K}_{i}^{*})\dot{\hat{K}}_{i}\nonumber\\
& =  r(- \gamma  r - \rho sat({r}) +\varphi) - \lambda e^2 +\sum_{i=0}^{1}(\hat{K}_{i} -{K}_{i}^{*})\dot{\hat{K}}_{i}.\label{vdot} 
\end{align}
Using the structure of $sat(\cdot)$ function from (\ref{input rob}), two scenarios can be identified:
\begin{itemize}
\item Scenario 1: $|r| \geq \epsilon$;
\item Scenario 2: $|r| < \epsilon$.
\end{itemize} 
Stability for both these scenarios are analysed subsequently.

\textbf{Scenario 1: $|r| \geq \epsilon$} 

The fact that $\hat{K}_i \geq 0$ from (\ref{low_bound}) yields $\rho \geq 0$. Utilizing this relation and the upper bound (\ref{psi_bound}), from (\ref{vdot}) we have
\begin{align}
&\dot{V} = r(- \gamma  r - \rho (r/|r|) + \varphi) - \lambda e^2 +\sum_{i=0}^{1}(\hat{K}_{i} -{K}_{i}^{*})\dot{\hat{K}}_{i} \nonumber\\
& \leq  - \gamma  r^2 - \lambda e^2 -\sum_{i=0}^{1} \left \lbrace (\hat{K}_{i} -{K}_{i}^{*})(||\xi || ^i | r |   - \dot{\hat{K}}_{i}) \right \rbrace. \label{part 2_m}
\end{align}
Form the adaptive laws (\ref{hat_k_0})-(\ref{hat_k_1}), the following equality holds
\begin{align}
  (\hat{K}_{i}-{K}_{i}^{*})\dot{\hat{K}}_{i} & = (\hat{K}_{i}-{K}_{i}^{*})( || \xi ||^{i} |{r} | - \alpha_{i}\hat{K}_{i}) \nonumber\\
& = (\hat{K}_{i} - K_{i}^{*})  || \xi ||^{i}|r| + \alpha_{i}\hat{K}_{i}{K}_{i}^{*} -\alpha_{i}\hat{K}_{i}^2.  \label{part 3_m} 
\end{align}
Substituting (\ref{part 3_m}) in (\ref{part 2_m}) yields
\begin{align}
\dot{V} \leq &   - \gamma r^2 - \lambda e^2 +\sum_{i=0}^{1} \left(\alpha_{i}\hat{K}_{i}{K}_{i}^{*} -\alpha_{i}\hat{K}_{i}^2 \right)\nonumber\\
 \leq &   - \gamma r^2 - \lambda e^2  -\sum_{i=0}^{1}  \left( \frac{\alpha_i}{2}(\hat{K}_{i}-{K}_{i}^{*})^2 - \frac{\alpha_i{K_{i}^{*}}^2}{2} \right) , \label{part 5_m}
\end{align}
where the terms under the summation are obtained based on the following simplification
\begin{align}
\hat{K}_{i}{K}_{i}^{*} -\hat{K}_{i}^2 =& -\left( \frac{\hat{K}_i}{\sqrt{2}}-\frac{{K}_i^{*}}{\sqrt{2}} \right)^2-\frac{\hat{K}_i^2}{2} + \frac{{K_{i}^{*}}^2}{2} \nonumber\\
\leq & -  \left( \frac{\hat{K}_i}{\sqrt{2}}-\frac{{K}_i^{*}}{\sqrt{2}} \right)^2 + \frac{{K_{i}^{*}}^2}{2}. \label{new_ineq}
\end{align}
Using the definition of Lyapunov function (\ref{lyap}) we have 
\begin{align}
V \leq  \max \lbrace J/2, 1/2 \rbrace \left( {r^2} +  e^2 + \sum_{i=0}^{1}(\hat{K}_{i} -{K}_{i}^{*})^2 \right). \label{lyap_up_boundm}
\end{align}
Using (\ref{lyap_up_boundm}), 
$\dot{V}$ in (\ref{part 5_m}) is further simplified to
\begin{align}
& \dot{V} \leq  -{\varrho} V+ \frac{1}{2} \sum_{i=0}^{1} \alpha_i{K_{i}^{*}}^2 , \label{part 6_m}
\end{align}
where $$ \varrho \triangleq \frac{\min_{ i} \lbrace \gamma, ~\lambda,~ \alpha_i/2 \rbrace} {\max \lbrace J/2,~ 1/2  \rbrace}  >0,~~i=0,1 $$
 by design (cf. (\ref{input rob}), (\ref{init})). 
Using $0< \kappa < {\varrho}$, $ \dot{V}$ in (\ref{part 6_m}) gets simplified to
\begin{align}
\dot{V} \leq &- \kappa V - ({\varrho} - \kappa)V + \frac{1}{2}\sum_{i=0}^{1} \alpha_i{K_{i}^{*}}^2. \label{part 7} 
\end{align}
\textbf{Scenario 2: $|r| < \epsilon$}

For this scenario using (\ref{input rob}) and (\ref{vdot}) we have
\begin{align}
\dot{V} & = r(- \gamma  r - \rho sat({r}) +\varphi) - \lambda e^2 +\sum_{i=0}^{2}(\hat{K}_{i} -{K}_{i}^{*})\dot{\hat{K}}_{i} \nonumber\\
& = r(- \gamma  r - \rho (r/\epsilon) + \varphi) - \lambda e^2 +\sum_{i=0}^{1}(\hat{K}_{i} -{K}_{i}^{*})\dot{\hat{K}}_{i} \nonumber\\
& \leq  - \gamma  r^2 - \lambda e^2 + \sum_{i=0}^{1} {K}_{i}^{*} ||\xi || ^i | r | +\sum_{i=0}^{1}  (\hat{K}_{i} -{K}_{i}^{*}) \dot{\hat{K}}_{i}. \label{part 2_m1}
\end{align}
Substituting (\ref{part 3_m}) into (\ref{part 2_m1}) and using (\ref{new_ineq}), $\dot{V}$ in (\ref{part 2_m1}) is simplified to
\begin{align}
\dot{V} & \leq   - \gamma r^2 - \lambda e^2 + \sum_{i=0}^{1} \hat{K}_{i} ||\xi || ^i | r | +\sum_{i=0}^{1} \left(\alpha_{i}\hat{K}_{i}{K}_{i}^{*} -\alpha_{i}\hat{K}_{i}^2 \right)\nonumber\\
& \leq    - \gamma r^2 - \lambda e^2 +  \sum_{i=0}^{1} \hat{K}_{i} ||\xi || ^i | r |  -\sum_{i=0}^{1}  \left( \frac{\alpha_i}{2}(\hat{K}_{i}-{K}_{i}^{*})^2 - \frac{\alpha_i{K_{i}^{*}}^2}{2} \right). \label{part 9}
\end{align}
Since $r$ is upper bounded by $\epsilon$ in this scenario, it implies $e, \dot{e} \in \mathcal{L}_{\infty}$ from (\ref{r}), thus $\xi \in \mathcal{L}_{\infty}$. It can be noted from the adaptive laws (\ref{hat_k_0})-(\ref{hat_k_1}) that $r, \xi \in \mathcal{L}_{\infty} \Rightarrow \hat{K}_0, \hat{K}_1 \in \mathcal{L}_{\infty}$. Therefore, $\exists \varsigma \in \mathbb{R}^{+}$ such that
\begin{align}
\sum_{i=0}^{1} \hat{K}_{i} ||\xi || ^i | r | \leq \varsigma. \label{var}
\end{align}
Then, following the similar lines which led to obtain (\ref{part 6_m}) in Scenario 1, we have the following for Scenario 2:
\begin{align}
 \dot{V} & \leq  -{\varrho} V+\varsigma+ \frac{1}{2} \sum_{i=0}^{1} \alpha_i{K_{i}^{*}}^2  \nonumber\\
 & = - \kappa V - ({\varrho} - \kappa)V +\varsigma+ \frac{1}{2} \sum_{i=0}^{1} \alpha_i{K_{i}^{*}}^2. \label{part 7_m1}
\end{align}
Let us define a scalar
\begin{align}
\mathcal{ B} \triangleq \frac{2\varsigma+\sum_{i=0}^{1} {\alpha_i{K_{i}^{*}}^2}}{ 2\left( \varrho - \kappa \right)}. \label{B} 
\end{align}
Observing the stability conditions (\ref{part 7}) and (\ref{part 7_m1}) for Scenarios 1 and 2 respectively, it can be seen that $\dot{V}(t) \leq - \kappa\ V(t) $ when $V(t) \geq \mathcal{B}$ so that 
\begin{align}
V \leq \max \left \lbrace V(0),  \mathcal{  B} \right \rbrace, ~~\forall t\geq 0, \label{ub_1m}
\end{align}
From the definition of the Lyapunov function (\ref{lyap}) one has $V \geq ({1}/{2}) |  e | ^2$, which in turn yields the ultimate bound $\omega$ as

\begin{align}
\omega \triangleq \sqrt{ \frac{2\varsigma+\sum_{i=0}^{1} {\alpha_i{K_{i}^{*}}^2}}{ \left( \varrho - \kappa \right)}},
\end{align}
which is global and uniform as it is independent of initial conditions.

\begin{remark}[The role of various gains and their selections]\label{select_d}
	It can be noted from (\ref{part 9}) that high values of $\gamma$ and $\lambda$ can improve convergence of the Lyapunov function and improve tracking accuracy, albeit at the cost of higher control input (cf. (\ref{r}) and (\ref{input rob})). On the other hand, the adaptive laws (\ref{hat_k_0})-(\ref{hat_k_1}) reveal that low values of $\alpha_0, \alpha_1$ lead to faster adaptation; however, it may cause high value of gains $\hat{K}_0, \hat{K}_1$ (i.e., over-estimation) and consequently high control input. Therefore, the gains $\gamma, \lambda, \hat{K}_0$ and $ \hat{K}_1$ should be selected in practice by maintaining a trade-off between performance and control input demand.
	 
\end{remark}

\section{Simulation Results}

\begin{figure}[!t]
	\centering
	\includegraphics[width=4in,height=2.5in]{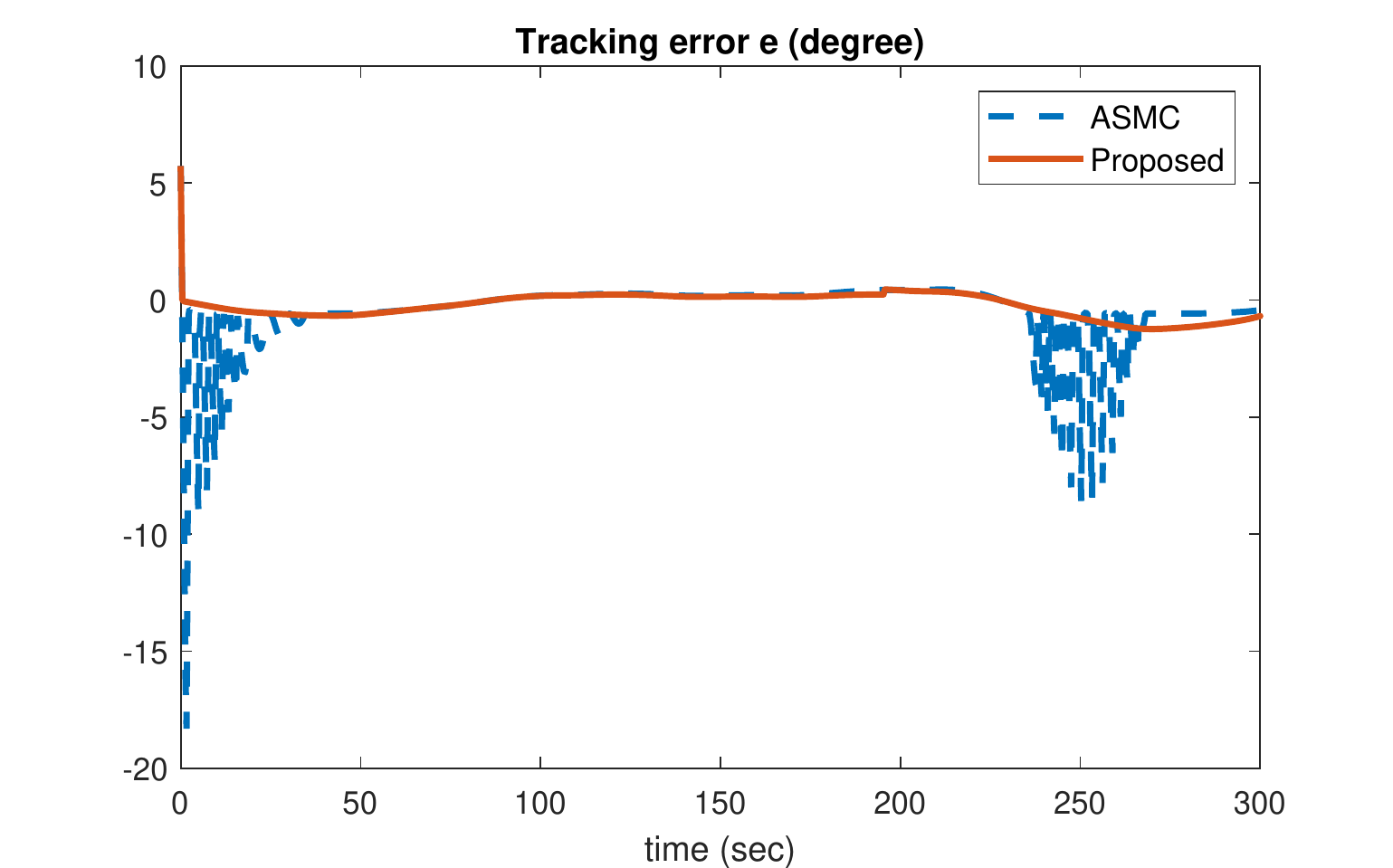}
	\caption{{Tracking performance comparison under sinusoidal steering (proposed controller with {$\lambda=100$)}.}}\label{fig:2} 
\end{figure}
The effectiveness of the proposed controller is verified compared to the {Adaptive Sliding Mode Control} (ASMC) method \cite{plestan2010new}, which does not require knowledge of system parameters, but considers the uncertainties to be bounded a priori by a constant. ASMC \cite{plestan2010new} follows the control law
\begin{align}
\tau &=-K sgn(s),\label{u_plestan}\\
~\dot{K}&= \begin{cases} \bar{K} | s | sgn(| s | - \epsilon)~~& \text{if}~K \geq \mu \\
\mu ~~& \text{if} ~K <\mu
\end{cases}, \label{plestan_law}
\end{align}
where $\bar{K}, \mu$ are user defined scalars selected as $\bar{K}=1, \mu=0.01$ for the simulations. To avoid chattering, the signum function in (\ref{u_plestan}) is replaced by the saturation function as in (\ref{input rob}) with $\epsilon=0.1$. 

To simulate a SBW system, the system dynamics parameters are selected as \cite{ATC}: $J=0.14;B=0.8;i_{rc}=8\times 10^{-3}$. In contrast to the conventional linear-in-parameters structure, we have selected a nonlinear-in-parameters dynamics by designing the friction force as $F=0.5 \tanh(\dot{\theta})+ \exp(-(\dot{\theta}/0.1)^2)$ where the first term approximates a Coulomb friction and the last term emulates Stribeck effect \cite{annaswamy1998adaptive}. The external disturbances are selected as $F_{rack}=1000\sin(0.03t); \tau_a=5\sin(0.05t)$. 

The control design parameters for the proposed controller are selected as: ${\lambda=100}, \gamma=20, \alpha_0=\alpha_1=0.1$. For parity, the sliding surface of ASMC is selected same as $r$ and the initial conditions are selected as $K(0)=\hat{K}_0(0)=\hat{K}_1(0)=0.001$, $\theta(0)=0.1, \dot{\theta}(0)=0$.

\begin{figure}[!h]
	\centering
	\includegraphics[width=5in,height=3.5in]{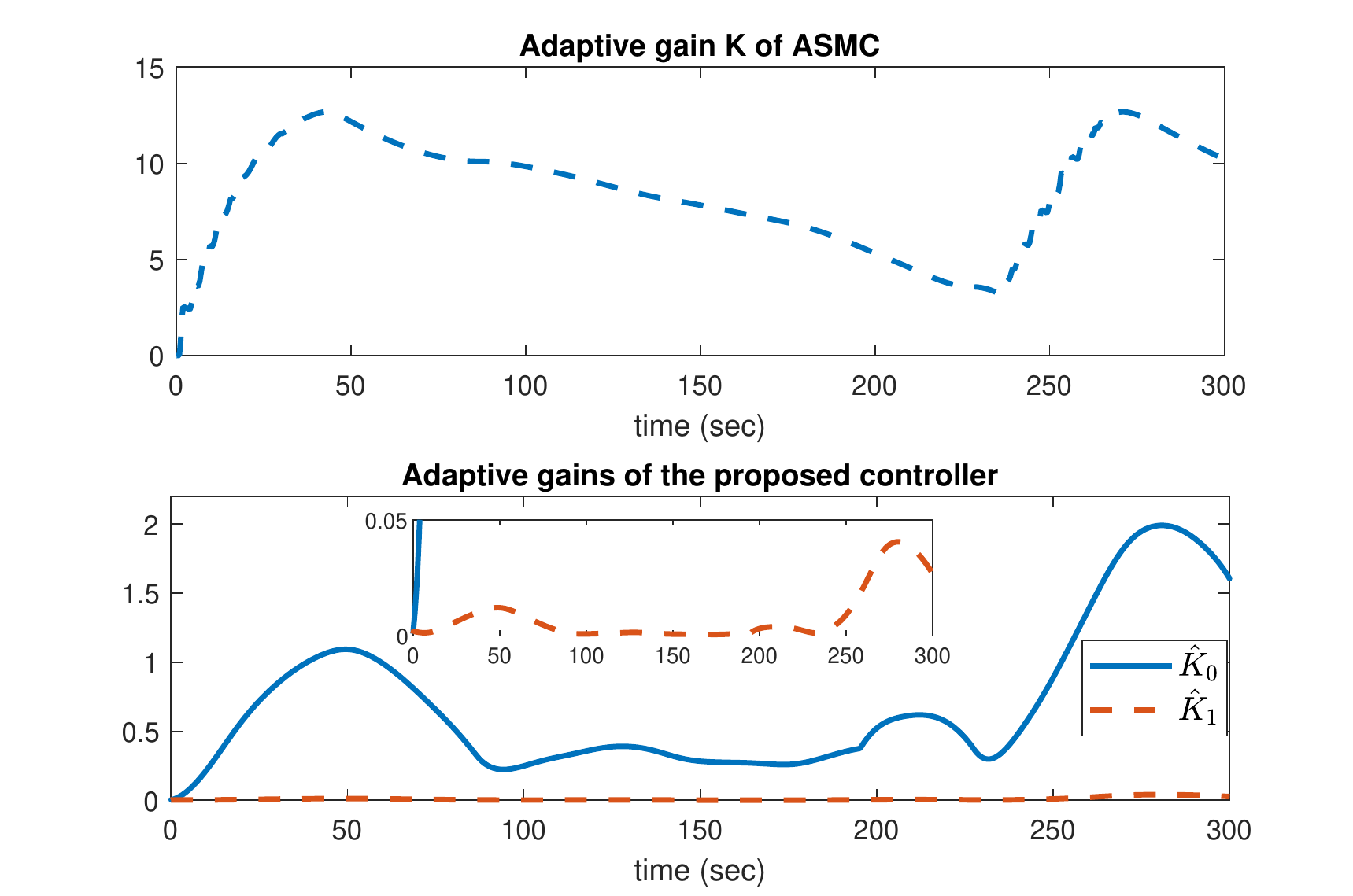}
	\caption{{Evolution of adaptive gains for various controllers.}}\label{fig:4} 
\end{figure}
\begin{figure}[!h]
	\centering
	\includegraphics[width=5in,height=2.8in]{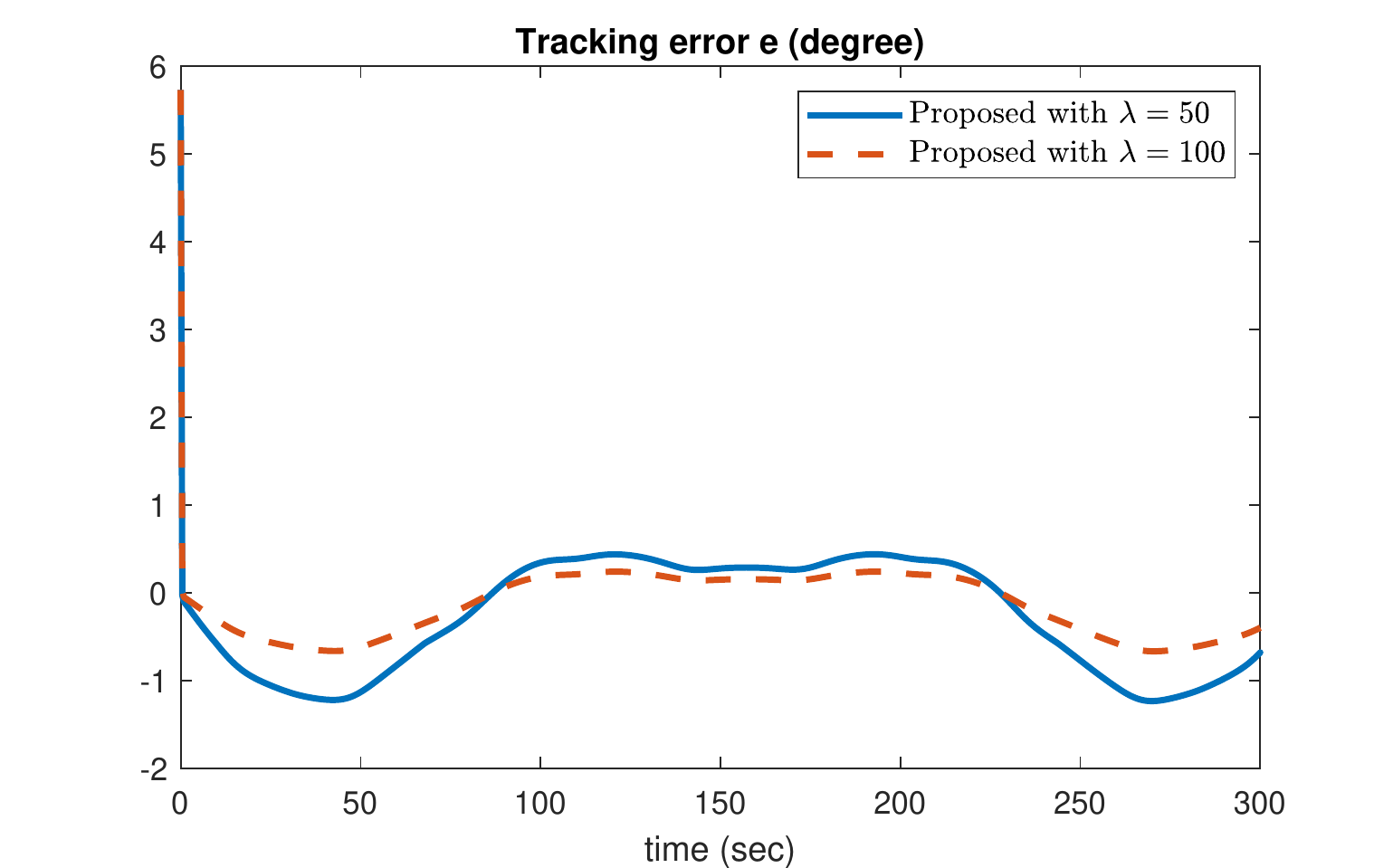}
	\caption{{Tracking performance comparison of the proposed controller with different control gains.}}\label{fig:5} 
\end{figure}

The tracking performance comparison of the proposed controller with that of ASMC as in Fig. \ref{fig:2} highlights the effectiveness of the proposed controller. Further, for better inference, the performance of various controllers are tabulated in Table \ref{table 2} in terms of root-mean-squared values of tracking error and of control input; the percentage values in the table show improvements over ASMC. Since ASMC relies on a priori bounded uncertainty, its adaptive gains (cf. Fig. \ref{fig:4}) are insufficient to tackle uncertainty even though it consumes higher control effort over the proposed design (cf. Table \ref{table 2}). Whereas, thanks to its state-dependent uncertainty based design, the proposed design provides significantly improved performance with lower adaptive gains and control effort.

\begin{table}[!h]
\renewcommand{\arraystretch}{1.3}
\caption{{Performance Comparison of Various Controllers}}
\label{table 2}
\centering
{
{{	\begin{tabular}{c c c c}
		\hline
		\hline
		 \multicolumn{4}{c}{RMS error (degree)} \\
		\cline{2-4}
		{Controller}& ASMC & Proposed with $\lambda=100$  & Proposed with $\lambda=50$  \\
		\hline
		 &  0.785 & 0.517 (34.14\%)  & 0.697 (11.21\%)   \\
		\hline
		 \multicolumn{4}{c}{RMS $\tau$ (Nm)} \\ \cline{2-4}
		 {Controller}& ASMC & Proposed with $\lambda=100$  & Proposed with $\lambda=50$  \\
		\hline
		 & 11.067  & 6.957 (37.13\%)& 6.196 (44.01\%) \\

		\hline
		\hline
\end{tabular}}}}
\end{table}

To study the trade-off between performance and control input required based on the choice design parameter, the simulation study is repeated for the proposed controller with $\lambda=50$ and, the performance comparison with $\lambda=100$ and $\lambda=50$ are given in Fig. \ref{fig:5} and Table \ref{table 2}. As discussed earlier (cf. Remark 5), these results demonstrate that with increased $\lambda$, tracking performance also improves but at the cost of higher control effort. Nevertheless, for both $\lambda=100$ and $\lambda=50$, the proposed controller outperforms ASMC.


\chapter{Adaptive Control of Steer-by-Wire Systems with Time Varying Input Delay and State-Dependent Uncertainty}
\label{ch:chap4}

In the previous chapter, the main focus was on tackling unknown state-dependent uncertainties in the SBW systems. However, the case of actuator delay was ignored. These delays are inherently encountered in many real-life scenarios \cite{Ref:1, Ref:3, ye2020switching, patel2021wide, ye2019observer}. Left uncompensated, time delays can cause the controllers to perform poorly which might reflect on the performance of the vehicle. 
Therefore, it is necessary such scenarios be included in the control design.

In view of the gaps in the state-of-the-art as discussed in Chapter 1 (cf. Sect. 1.4.2), this work proposes a new adaptive-robust control law named, Adaptive-Robust Time Delay Control (ARTDC) for SBW systems which is subjected to arbitrarily varying input delay and state-dependent uncertainty. The highlights of the proposed scheme, especially compared to \cite{Ref:self}, are listed below: 
\begin{itemize}
\item ARTDC avoids any a priori assumption of uncertainties being upper bounded by a constant. This enhances the applicability of the proposed scheme. 
\item Under similar selection
of \textit{nominal control parameters}\footnote{here nominal control parameters signify the gains which are not designated to tackle uncertainties. }, ARTDC enhances the maximum allowable input delay for the system compared to \cite{Ref:self} (cf. Lemma 4), while still remaining insensitive towards variations in input delay.

\end{itemize}

The effectiveness of ARTDC for SBW systems is verified via simulation.

The remainder of this chapter is organized as follows: the control problem is highlighted in Section 3.1; Section 3.2 details the proposed control scheme along with rigorous closed-loop stability analysis. This section also presents some key
advantages of the proposed ARTDC law compared to the state of the art; Sections 3.3 and 3.4 present the simulation results.

\section{Problem Formulation}
The SBW system dynamics as in (\ref{dyn})

\begin{equation}
    J\ddot{\theta}(t) + B\dot{\theta}(t) + F (t) + i_{rc}F_{rack} + \tau_a = \tau(t), 
\end{equation}
can be rewritten after incorporating the input delay as
\begin{equation}\label{sys}
\ddot{\theta}(t)=f(\dot{\theta}(t))+g\tau(t-h_{in}),
\end{equation}
where $f=-J^{-1}( B\dot{\theta} + F + i_{rc}F_{rack} + \tau_a)$, $g=J^{-1}$ and $h_{in} (t)$ is a time varying input delay. The following assumptions define the uncertainties present in system (\mathbbf{\ref{sys}}):
\begin{assum}\label{prop}
Let the dynamics term $f$ can be decomposed as $f= \hat{f} + \Delta f$ where $\hat{f}$ is the known (nominal) part and $\Delta f$ is the unknown part having the following upper bound structure
\begin{equation}
| \Delta f (\dot{\theta})| \leq \gamma^{**}_0+\gamma^{**}_1 ||x|| \label{f_prop}
\end{equation}
where $\gamma^{**}_0, \gamma^{**}_1 \in \mathbb{R}^{+}$ are unknown finite constants and $x = [\theta,~\dot{\theta}]$.
\end{assum}
\begin{assum}
The nominal part of $g$, i.e., $\hat{g}$ can be designed in a way such that
\begin{equation}
 | (g\hat{g}^{-1}-1) | \triangleq | \bar{g} | < 1 ~~\forall h_{in}. \label{mass}
\end{equation}
\end{assum}
\begin{assum}
There exists a known scalar $\bar{h}_{in}$ such that $|h_{in}(t) | \leq \bar{h}_{in}$. However, instantaneous values of delay, $h_{in}(t)$, and its time derivatives along with their bounds are unknown.
\end{assum}
\begin{remark}
Assumption 3.1 can be verified from (\ref{psi_bound}) in Chapter 2, while Assumption 3.2 is quite common for practical electro-mechanical systems, stemming from the physical laws that govern the dynamics (cf. \cite{roy2019simultaneous}). On the other hand, Assumption 3.3 defines that the input delay can have unknown (arbitrary) variation, i.e., no restriction is imposed on the rate of variation in $h_{in}$.
\end{remark}
Let there exist constant matrices ${P}\in \mathbb{R}^{2 \times 2}>0$ and ${Q}\in \mathbb{R}^{2 \times 2}>0$, and scalar $K>0$ such that the following conditions hold:
 (i) ${P}=\begin{bmatrix}
{P}_1 & {P}_2 \\ 
{P}_2^{T} & {P}_3
\end{bmatrix}$, where scalars ${P}_1>0$, ${P}_2>0$, ${P}_3>0$, ${P}_{3}^{-1}{P}_2>0$,\\
(ii) ${A}=\begin{bmatrix}
0 & 1 \\ 
-K & -2\Omega 
\end{bmatrix}$ is Hurwitz, where $\Omega ={P}_{3}^{-1}{P}_2>0$,\\
 (iii) ${A}^T{P}+{P}{A}=-{Q}$.\\
In \cite{Ref:self}, the matrix $P$ is defined as the solution of the following Lypaunov equation 
\begin{equation}
\bar{A}^T P+P\bar{A}=-Q \label{ly}
\end{equation}
where $\bar{A}=\begin{bmatrix}
0 & 1 \\ 
-K_1 & -K_2
\end{bmatrix}$, $K_1>0, K_2>0$. So, the values of $P$ and $Q$ in conditions (i) and (iii) will be similar to that of \cite{Ref:self} if
\begin{equation}
K_1=K \quad  \text{and} \quad  K_2=2 \Omega. \label{comp}
\end{equation}
Establishment of relation (\mathbbf{\ref{comp}}) is necessary for the comparison between the proposed controller and the work \cite{Ref:self} as discussed later (cf. Lemma 4).


\textbf{Control Problem:} Under Assumptions 1-3, design an adaptive-robust control framework for system (\ref{sys}) to track a desired trajectory $\theta^d(t)$ where the selection of desired trajectory satisfies $\theta^d, \dot{\theta}^d, \ddot{\theta}^d \in \mathcal{L}_{\infty}$. 

\section{Adaptive-Robust Time Delay Controller (ARTDC): Design and Analysis}
Let $e(t)=\theta(t)-\theta^d(t)$ be the tracking error and $\xi=[e^T \quad \dot{e}^T]^T$. The control input $\tau$ for ARTDC is defined as
\begin{equation}\label{input1}
\tau=\hat{g}^{-1}(\hat{u}+\Delta u- \hat{f}),
\end{equation}
where $\hat{u}$ is the nominal control designed as
\begin{align}
&\hat{{u}}=\ddot{\theta}^d-\Omega\dot{e}\label{aux2} 
\end{align}
and $\Delta u$ is the switching control input responsible to negotiate the uncertainties (defined later). 

Further, $\tau(t)$ goes through an input delay of $h_{in}$ as
\begin{align}
\tau(t-h_{in})&=\hat{g}^{-1}(t-h_{in})\lbrace \hat{{u}}(t-h_{in})+\Delta {u}(t-h_{in})  - \hat{f}(t-{in})\rbrace \label{input}
\end{align}
Taking $B=\begin{bmatrix}
0&
1
\end{bmatrix}^T$, let us define an error variable as
\begin{align}
s&=B^TP\xi= {P}_3 \dot{e}+{P}_2 e\nonumber\\
\Rightarrow {P}_3^{-1}s &=\dot{e}+\Omega e.\label{s}
\end{align}
Time derivative of (\mathbbf{\ref{s}}) yields
\begin{align}\label{s dot}
 {P}_3^{-1} \dot{s} &= \ddot{e}+ \Omega \dot{e}\nonumber\\
 &=\ddot{\theta}-\ddot{\theta}^d+ \Omega \dot{e}\nonumber\\
 &=f+g\tau(t-h_{in})-\ddot{\theta}^d+ \Omega \dot{e}.
\end{align}
Substituting (\mathbbf{\ref{input}}) into (\mathbbf{\ref{s dot}}) we get,
\begin{align}
{P}_3^{-1}\dot{s} &= f-\hat{f}(t-h_{in})+\Delta {u}(t-h_{in})+\hat{{u}}(t-h_{in})\nonumber\\
&\qquad +(g\hat{g}^{-1}-1)\tau(t-h_{in}) -\ddot{\theta}^d+ \Omega \dot{e}. \label{s dot new}
\end{align}
Further, taking time derivative of (\mathbbf{\ref{s}}) and utilizing (\mathbbf{\ref{s dot new}}) yields the following:
\begin{align}
\ddot{e}&=-Ke-\Omega \dot{e}+{P}_3^{-1}\dot{s}+Ke\nonumber\\
&=-Ke-\Omega \dot{e}-\Omega \dot{e}(t-h_{in})+\Delta {u}(t-h_{in})+\sigma+Ke+\Omega \dot{e}+\bar{g}\Delta {u}(t-h_{in})\label{error dyn}
\end{align}
 where $\bar{g} = (g\hat{g}^{-1}-1)$ and
 \begin{equation}
  \sigma=f-\hat{f}(t-h_{in})+\bar{g}(\hat{u}(t-h_{in}) - \hat{f}(t-h_{in}))  -\ddot{\theta}^d+\ddot{\theta}^d(t-h_{in})
\end{equation} 
 is defined as the \textit{overall uncertainty}. 

\begin{lemma}[\cite{roy2019reduced}]\label{assum 3}
Under the assumption of the system (\mathbbf{\ref{sys}}) being locally Lipschitz continuous for all delay, the upper bound of $\sigma$ follows the structure 
\begin{equation}
|\sigma| \leq \gamma^{*}_0+\gamma^{*}_1 ||\xi|| \label{n sigma}
\end{equation}
where $\gamma^{*}_0, \gamma^{*}_1 \in \mathbb{R}^{+}$ are unknown finite constants.
\end{lemma}
\begin{remark}
In \cite{Ref:self}, $\sigma$ is assumed to be upper bounded by an unknown constant: such assumption is conservative in nature because, due to the involvement of states in the upper bound of $|\sigma|$ as in (\mathbbf{\ref{n sigma}}), it puts a priori restriction on the system states. Similar assumption is commonly found in the literature of (non-delayed) adaptive-robust designs (cf. the discussions as in \cite{roy2019simultaneous, roy2019, roy2019auto}). In the following, we propose a new framework avoiding such assumption.
\end{remark}

After defining the structure of $|\sigma|$ as in (\mathbbf{\ref{n sigma}}), we now design $\Delta {u}$ as  
\begin{align}
&\Delta {u}(t) =  -\zeta (t) \text{sgn}(s(t)) , ~\zeta (t)  =  \frac{1}{(1-|\bar{g}|)}\lbrace c(t)+\beta(t)+\rho(t) \rbrace,\label{delta u2}\\
  &c(t)=\hat{\gamma}_0(t)+\hat{\gamma}_2(t)+\hat{\gamma}_1(t)||\xi|| \label{c}
\end{align}
where the switching gains $\hat{\gamma}_0,\hat{\gamma}_1 \in \mathbb{R}^{+}$ are responsible to tackle uncertainties while the other switching gains $\beta,\rho,\hat{\gamma}_2 \in \mathbb{R}^{+}$ play crucial role in closed-loop stability (cf. Remark {\ref{rho_beta}}). The switching gains are evaluated as
\begin{align}
 \dot{\hat{\gamma}}_j & =
  \begin{cases}
    -{\alpha}_j ||\xi||^j | s |       &  \text{if } s \dot{s} \leq 0 \lor \beta \leq \underline{\beta}\lor \rho \leq \underline{\rho} \\
   {\alpha}_j ||\xi||^j | s |       & \text{if }  \hat{\gamma}_i \leq \underline{\gamma}_i \lor  s \dot{s} >0 
  \end{cases}\label{alpha0}\\
   \dot{\hat{\gamma}}_2 & =
  \begin{cases}
    -\varsigma \alpha_2 ||\xi||^3       &  \text{if } s \dot{s} \leq 0 \lor \beta \leq \underline{\beta} \lor \rho \leq \underline{\rho}\\
   {\alpha}_2 ||\xi|| | s |      & \text{if }  \hat{\gamma}_i \leq \underline{\gamma}_i \lor  s \dot{s} >0 
  \end{cases}\label{alpha1}\\
  \dot{\beta} & =
  \begin{cases}
    -\frac{1}{\beta}       & \quad \text{if } \beta > \underline{\beta}\\
    \delta     & \quad \text{if }  \beta \leq \underline{\beta}
  \end{cases}, 
  \dot{\rho}  =
  \begin{cases}
    -\frac{|s|}{\rho}       & \quad \text{if } \rho > \underline{\rho}\\
    \delta |s|       & \quad \text{if }  \rho \leq \underline{\rho}
  \end{cases}\label{beta1}\\
   &j=0,1, \hat{\gamma}_{i0}   > \underline{\gamma}_i,, i=0,1,2, \rho_0  > \underline{\rho}, \beta_0  > \underline{\beta} \nonumber
\end{align}
where $\hat{\gamma}_{i0}, \rho_0, \beta_0$ are the initial values of $\hat{\gamma}_{i},\rho, \beta$ respectively and $\underline{\gamma}_i, \underline{\beta}, \underline{\rho}, {\alpha}_i, \varsigma, \delta \in \Re^{+}$ are user-defined scalars.
 
From (\mathbbf{\ref{alpha0}})-(\mathbbf{\ref{beta1}}) it can be inferred that
\begin{equation} \label{bound}
\hat{\gamma}_{i}(t)   \geq \underline{\gamma}_i,\quad \forall i=0,1,2, \quad \underline{\rho} \leq \rho(t) \leq \rho_0 \quad \text{and} \quad \underline{\beta} \leq \beta(t) \leq \beta_0 \quad \forall t.
\end{equation}

\section{Stability Analysis}
Before presenting the closed-loop stability result, the following Razumikhin lemma is stated which is instrumental in deriving a tolerance of time delay for the system:
\begin{lemma}\label{assum 4}
Using the Razumikhin-type theorem \cite{Ref:36}, let the following standard inequality holds for a Lyapunov function candidate $ V(\xi)$:
\begin{equation}\label{razu}
V(\xi(\nu ))<rV(\xi(t)),\qquad t-2h_{in}\leq \nu \leq t.
\end{equation}
where $r>1$ is a constant.
\end{lemma} 
\begin{remark}
The condition (\mathbbf{\ref{razu}}) should not be considered as an assumption. On the contrary, it represents an analytical condition. Interested readers may see \cite[Chap. 1]{roy2019book} for a detailed discussion on this.
\end{remark}
Stability is analysed using the following Lyapunov function candidate:
\begin{align}
V&=V_1+\sum_{i=0}^{1}\frac{1}{2{\alpha}_i}(\hat{\gamma}_i-\gamma^{*}_i)^2 +\frac{1}{2 \alpha_2} \hat{\gamma}^2_2 + \frac{1}{2}\Gamma \beta^2+\frac{1}{2}\Gamma_1 \rho^2 \label{lyapunov}
\end{align}
where $V_1=\frac{1}{2}\xi^TP\xi$ and the terms $\Gamma,\Gamma_1 \in \mathbb{R}^{+}$ would be defined later.

Before inspecting the closed-loop system stability, two important lemmas are stated: Lemma 3.3 simplifies $\dot{V}_1$ which consequently reveals the maximum allowable input delay $\bar{h}_{in}$ and Lemma 3.4 identifies the condition under which the proposed work can extend the allowable delay over the state of the art.

\begin{lemma}
(i) $\dot{V}_1(\xi)$ can be expressed in a simplified way as
\begin{align}
\dot{V}_1(\xi) &\leq -\frac{1}{2}\xi^{T}\left ( Q-hG \right )\xi+ \Gamma+||s||\Gamma_1 + s(\Delta u+\sigma+(PB)^{-1}PC \xi + \bar{g}\Delta {u}). \label{delay cal 1}
\end{align}
(ii) The upper bound of delay i.e. $\bar{h}_{in}$ for system (\mathbbf{\ref{sys}}) is found to be
\begin{equation}\label{delay value}
h_{in}< \frac{\lambda _{\min}(Q)}{|| G ||}:=\bar{h}_{in},
\end{equation}
where $$G =( \eta PB_1( A_1P^{-1}A_1^T+B_1P^{-1}B_1^{T}+P^{-1})B_1^{T}P+2(r/\eta )P );$$ the matrices $P$, $K$, $Q$ (cf. (\mathbbf{\ref{ly}}), (\mathbbf{\ref{comp}})) and scalar design parameters $r>1, \eta>0$ are selected in a manner which satisfies 
\begin{equation*}
\lambda _{min}(Q)>h_{in}|| G || \quad \forall h_{in}.
\end{equation*}

\end{lemma}
\textit{Proof.}
The error dynamics (\mathbbf{\ref{error dyn}}) can be written in the sate-space form as
\begin{equation}\label{error dyn state}
\dot{\xi}=A_1 \xi+B_1 \xi (t-h_{in})+B(\sigma+\Delta {u}(t-h_{in})+\bar{g}\Delta {u}(t-h_{in}))+C\xi
\end{equation}
where $A_1=\begin{bmatrix}
0 & 1 \\
-K & -\Omega
\end{bmatrix} 
, B_1=\begin{bmatrix}
0 & 0 \\
0 & -\Omega
\end{bmatrix}, C=\begin{bmatrix}
0 & 0 \\
K & \Omega
\end{bmatrix}$.

\noindent Further, $\xi(t-h_{in})=\xi(t)-\int\limits_{-h_{in}}^0 \dot{\xi}(t+\psi)\mathrm{d}\psi$, where the derivative inside the integral is with respect to $\psi$. So, (\mathbbf{\ref{error dyn state}}) can alternatively be expressed as
\begin{align}
\dot{\xi}(t)&=A \xi(t)-B_1\int\limits_{-h_{in}}^0 \dot{\xi}(t+\psi)\mathrm{d}\psi+C\xi  +B(\sigma+\Delta {u}(t-h_{in})+\bar{g}\Delta {u}(t-h_{in})) . \label{error dyn delayed}
\end{align}
(i) Since $\xi^TPC\xi= s^T (PB)^{-1}PC \xi$ and using (\mathbbf{\ref{error dyn delayed}}) we have
\begin{align}
\dot{V}_1=&-\frac{1}{2}\xi^TQ \xi-\int_{-h}^{0}\xi^{T}PB_1[ A_c \xi(t+\psi )+ B_1\xi(t-h_{in}+\psi ) \nonumber\\
& +B\Phi (t+\psi) ]d\psi +s\lbrace(1+\bar{g})\Delta {u} (t-h_{in})+\sigma+(PB)^{-1}PC \xi\rbrace \label{lya_dot for time delay}
\end{align}
where $A_C=A_1+C$, $\Phi(t)=(1+\bar{g})\Delta u (t-h_{in})+\sigma(t)$.

Applying (\mathbbf{\ref{razu}}) to (\mathbbf{\ref{lyapunov}}), the following relation is achieved,
\begin{equation}
\xi^{T}(\nu )P \xi(\nu )<r\xi^{T}(t)P\xi(t)+\varphi(\nu), \label{new}
\end{equation}  
\begin{align}
\varphi(\nu)=&r\lbrace\sum_{i=0}^{1}\frac{1}{2{\alpha}_i}(\hat{\gamma}_i(t)-\gamma^{*}_i)^2 +\frac{1}{2 \alpha_2} \hat{\gamma}(t)^2_2+\Gamma\beta^2(t) +\Gamma_1 \rho^2(t)\rbrace \nonumber \\
& -\lbrace\sum_{i=0}^{1}\frac{1}{2{\alpha}_i}(\hat{\gamma}_i(\nu)-\gamma^{*}_i)^2 +\frac{1}{2 \alpha_2} \hat{\gamma}(\nu)^2_2+\Gamma \beta^2(\nu) +\Gamma_1 \rho^2(\nu)\rbrace.
\end{align}
For any two non-zero vectors $z$ and $\acute{z}$, there exist a constant $\eta>0$ and a matrix $D>0$ such that the following inequality holds:
\begin{equation}\label{ineq 1}
-2z^{T}\acute{z}\leq \eta z^{T}D^{-1}z+(1/\eta )\acute{z}^{T}D\acute{z}.
\end{equation}
Then, applying (\mathbbf{\ref{ineq 1}}) and taking $D=P$ the following inequalities are obtained
\begin{align}\label{cond1}
- 2\int_{-h_{in}}^{0}\xi^{T}PB_1 A_c\left [ \xi(t+\psi)\right ]d\psi \leq & \bar{h}_{in}\xi^{T} [ \eta PB_1A_cP^{-1}   A_c^{T}B_1^{T}P +(r/\eta )P)  ]\xi  \nonumber\\
&  +\int_{-h_{in}}^{0}(1/\eta)\varphi(t+\psi) d\psi.
\end{align}

\begin{align}\label{cond2}
- 2\int_{-h_{in}}^{0}\xi{T}PB_1B_1   \xi(t-h+\psi)d\psi \leq & \bar{h}_{in} \xi^{T} [ \eta PB_1B_1P^{-1} B_1^{T} B_1^{T} P+(r/\eta )P) ] \xi  \nonumber\\
&  +\int_{-h_{in}}^{0}(1/\eta)\varphi(t-h_{in}+\psi) d\psi .
\end{align}
\begin{align}\label{cond3}
- 2\int_{-h_{in}}^{0}\xi^{T}PB_1 \left [ B \Phi(t+\psi)\right ]d\psi \leq & \bar{h}_{in} \xi ^{T}\left [ \eta PB_1P^{-1}B_1^{T}P \right ]\xi\nonumber\\
& +\int_{-h_{in}}^{0}(1/\eta )B ^{T} \Phi(t+\psi)PB \Phi(t+\psi)  d\psi. 
\end{align}
Since (\mathbbf{\ref{alpha0}})-(\mathbbf{\ref{beta1}}) are piecewise continuous and assuming the system to be locally Lipschitz $\forall h_{in} \in [ \begin{matrix}
0 & \bar{h}_{in}
\end{matrix} ]$, then $\exists \Gamma, \Gamma_1 \in \Re^{+}$ such that the following inequalities hold:
\begin{align}
\frac{1}{2\eta }|| \int_{-h_{in}}^{0} & [ \varphi(t+\psi)+\varphi(t-h_{in}+\psi )\nonumber\\
&+(B \Phi(t+\psi) ^{T}PB \Phi(t+\psi) ]d\psi  || \leq \Gamma. \label{cond4}
\end{align}
\begin{align}
||(\Delta u-\Delta u(t-h_{in})) + (\bar{g}\Delta u- \bar{g}(t-h_{in}))\Delta u(t-h_{in})) ||  \leq \Gamma_1. \label{cond5}
\end{align} 
The conditions (\ref{cond4})-(\ref{cond5}) essentially imply that the rate of change in uncertainty is not faster than that of the input delay, which holds in practice for actuator delays \cite{roy2020time}.

Substituting (\ref{cond1})-(\ref{cond5}) into (\ref{lya_dot for time delay}), we have 
\begin{align}
\dot{V}_1(\xi) &\leq -\frac{1}{2}\xi^{T}\left ( Q-h_{in}G \right )\xi+ \Gamma+||s||\Gamma_1  + s^{T}(\Delta u+\sigma+(PB)^{-1}PC \xi + \bar{g}\Delta u). \label{delay cal}
\end{align}
(ii) For the closed-loop stability of the system, the first term of (\mathbbf{\ref{delay cal}}) is required to be negative, i.e.
\begin{equation*}
Q-h_{in}G>0 ~ \Rightarrow \lambda _{\min}(Q)>h_{in}|| G || \quad \forall h_{in}.
\end{equation*}
Hence, the maximum allowable delay is obtained as
\begin{equation}
h_{in}< \lambda _{\min}(Q)/|| G ||:= \bar{h}_{in}.\label{max1}
\end{equation}
\begin{remark}
The condition (\mathbbf{\ref{max1}}) implies that the parameters $Q, G$ are to be designed from an a priori given admissible/tolerable upper bound of input delay (cf. Assumption 3.3).
\end{remark}
After obtaining an allowable range of delay, the following lemma highlights how ARTDC is able to extend this range over the existing adaptive-robust design \cite{Ref:self}. 
\begin{lemma}
For similar choice of the matrices $P$ and $Q$ (when (\mathbbf{\ref{comp}}) is satisfied), ARTDC provides larger maximum allowable delay than AROLC as in \cite{Ref:self}, if the controller gains are designed to satisfy the following condition
\begin{equation}
\bar{P}_3>|\bar{P}_2 K \Omega^{-1}| ~\text{and}~ \bar{P}_1+\bar{P}_3> |\bar{P}_2|
\end{equation}
where
$$ P^{-1}=\begin{bmatrix}
\bar{P}_1 & \bar{P}_2\\ 
\bar{P}_2^T & \bar{P}_3
\end{bmatrix}$$.
\end{lemma}
\thextit{Proof.}
The maximum allowable delay provided by AROLC \cite{Ref:self} is given as:
\begin{equation}
h_{in}< \lambda _{\min}(Q)/|| G_1 ||:= \hat{h}_{in} \label{max2}
\end{equation}
where $G_1 =( \eta P \bar{B}_1( \bar{A}_1P^{-1}\bar{A}_1^T+\bar{B}_1P^{-1}\bar{B}_1^{T}+{P}^{-1})\bar{B}_1^{T}P+2(r/\eta )P )$, $\bar{A}_1=\begin{bmatrix}
0 & 1 \\
0 & 0
\end{bmatrix} 
, \bar{B}_1=\begin{bmatrix}
0 & 0 \\
-K_1 & -K_2
\end{bmatrix}$.\\

 Then, using (\mathbbf{\ref{comp}}), matrices $G$ (as in (\mathbbf{\ref{delay value}})) and $G_1$ (as in (\mathbbf{\ref{max2}})) are simplified as
\begin{align}
&G=\eta P\begin{bmatrix}
0 & 0 \\ 
0 & J
\end{bmatrix}P+2(r/\eta )P, G_1=\eta P\begin{bmatrix}
0 & 0 \\ 
0 & J_1
\end{bmatrix}P+2(r/\eta )P \label{g}\\
&J=(\Omega K \bar{P}_1 +2\Omega^2\bar{P}_2) K \Omega+2 \Omega^2 \bar{P}_3 \Omega^2+ \Omega \bar{P}_3 \Omega, \label{j1}\\
&J_1=(4\Omega K \bar{P}_1+16\Omega^2\bar{P}_2) K \Omega+16\Omega^2 \bar{P}_3 \Omega^2+4\Omega \bar{P}_3 \Omega,\nonumber\\
& \qquad \qquad +K(\bar{P}_3+\bar{P}_1)K+(K+2\Omega)\bar{P}_2 K. \label{j2}
\end{align}
Inspecting (\mathbbf{\ref{max1}}), (\mathbbf{\ref{max2}}) and (\mathbbf{\ref{g}}) it can noted that $\bar{h}_{in}>\hat{h}_{in}$ would be achieved if $J<J_1$. This implies that the following condition needs to be satisfied:
\begin{align}
&K(\bar{P}_3+\bar{P}_1)K+14\Omega^2 \bar{P}_3 \Omega^2+3\Omega \bar{P}_3 \Omega>-14\Omega^2\bar{P}_2 K \Omega \nonumber\\
&\qquad \qquad \qquad \qquad-K\bar{P}_2 K-2\Omega\bar{P}_2 K. \label{l1}
\end{align} 
On the other hand, the above condition will be satisfied if
\begin{align}
(i)~&\Omega^2 \bar{P}_3 \Omega^2>-\Omega^2\bar{P}_2 K \Omega\Rightarrow \bar{P}_3 \Omega>-\bar{P}_2 K, \label{l2}\\
(ii)~ &3\Omega \bar{P}_3 \Omega>-2\Omega\bar{P}_2 K\Rightarrow\bar{P}_3 \Omega>-(2/3)\bar{P}_2 K,\label{l3}\\
(iii)~& K(\bar{P}_3+\bar{P}_1)K>-K\bar{P}_2 K \Rightarrow\bar{P}_3+\bar{P}_1>-\bar{P}_2. \label{l4}
\end{align}
The condition (\mathbbf{\ref{l4}}) will hold if
\begin{equation}
\bar{P}_1+\bar{P}_3> |\bar{P}_2|.   \label{l5}
\end{equation}
Hence, taking into account the conditions (\mathbbf{\ref{l2}})-(\mathbbf{\ref{l3}}), $\bar{h}_{in}>\hat{h}_{in}$ can be achieved if
\begin{equation}
\bar{P}_3>|\bar{P}_2 K \Omega^{-1}|~\text{and}~\bar{P}_1+\bar{P}_3> |\bar{P}_2|.
\end{equation}

\begin{theorem}
The system (\mathbbf{\ref{sys}}) employing ARTDC with the control input (\mathbbf{\ref{input1}})-(\mathbbf{\ref{beta1}}) is asymptotically converging in tracking error $\xi$ when $\lbrace \hat{\gamma}_i \leq \underline{\gamma}_i \lor  s \dot{s} >0 , i=0,1,2\rbrace $ and UUB (Uniformly Ultimately Bounded) otherwise. 
\end{theorem}

\textit{Proof.} Exploring the various combinations of $\Delta u$ and gains $\hat{\gamma}_i, \beta, \rho$ in (\mathbbf{\ref{delta u2}}) and (\mathbbf{\ref{alpha0}})-(\mathbbf{\ref{beta1}}) respectively, the following two possible cases are identified:

\par \textbf{Case (i):}   $ \hat{\gamma}_i \leq \underline{\gamma}_i \lor  s \dot{s} >0 , i=0,1,2$
\par \textbf{Case (ii):} $  s \dot{s} \leq 0 \lor \beta \leq \underline{\beta}\lor \rho \leq \underline{\rho}$

Note that $|| \sigma + (PB)^{-1}PC \xi ||$ satisfies the upper bound structure (\mathbbf{\ref{n sigma}}) with the constant $|| (PB)^{-1}PC ||  $ getting added to $\gamma_1^*$. However, for convenience, we will use $\gamma_1^*= \gamma_1^{*'} + || (PB)^{-1}PC || $ with some abuse of notation, where $\gamma_1^{*'}$ is given in (\mathbbf{\ref{n sigma}}).

Utilizing Lemma 3.3 and (\mathbbf{\ref{delta u2}})-(\mathbbf{\ref{beta1}}), stability of the closed-loop system for various cases are investigated as follows:
\par \textbf{Case (i):} $  \hat{\gamma}_i \leq \underline{\gamma}_i \lor  s \dot{s} >0, i=0,1,2 $\\
Taking $\Upsilon=\left ( Q-hG \right )$, the time derivative of (\mathbbf{\ref{lyapunov}}) yields
\begin{align}
\dot{V} &\leq -\frac{1}{2}\xi^{T}\Upsilon\xi+ \Gamma+ s^{T} \lbrace-(c+\rho+\beta)\frac{s}{| s|}+\sigma \rbrace +|s|\Gamma_1 \nonumber\\
& \qquad \qquad + \sum_{i=0}^{1}(\hat{\gamma}_i-\gamma^{*}_i)||\xi||^i |s| + \frac{1}{\alpha_2}\hat{\gamma}_2 \dot{\hat{\gamma}}_2+ \Gamma \beta \dot{\beta}+\Gamma_1 \rho \dot{\rho} \nonumber\\
& \leq -\frac{1}{2} \lambda_{\min}(\Upsilon)||\xi||^2 \leq 0   \label{case 1}
\end{align}
\noindent From (\mathbbf{\ref{case 1}}) it can be inferred that ${V}(t) \in \mathcal{L}_{\infty}$. Thus, from the definition of $V$ we have $\xi(t), \hat{\gamma}_i(t), \rho(t), \beta(t) \in \mathcal{L}_{\infty}$ $\forall i=0,1,2$. This further implies $\tau(t), \sigma(t) \in \mathcal{L}_{\infty}$. Since, system follows locally Lipschitz condition $\forall h_{in} \in [0 \quad \bar{h}_{in}]$, from (\mathbbf{\ref{error dyn delayed}}) we have $\dot{\xi}(t) \in \mathcal{L}_{\infty}$. The boundedness of $\xi(t), \dot{\xi}(t)$ ensures $\ddot{V}(t) \in \mathcal{L}_{\infty}$ from (\mathbbf{\ref{case 1}}). Thus, using the Lyapunov like lemma \cite{Ref:khalil} yields
\begin{align*}
\lim_{t \to \infty} \dot{V}(t)=0 \Rightarrow \lim_{t \to \infty} \xi(t)=0 .
\end{align*}

\par \textbf{Case (ii):} $ s^T \dot{s} \leq 0 \lor \beta \leq \underline{\beta} \lor \rho \leq \underline{\rho} $\\
Using the fact $|s| \leq ||B^T P|| ||\xi||, \hat{\gamma}_2\geq \underline{\gamma}_2$ and (\mathbbf{\ref{delta u2}})-(\mathbbf{\ref{beta1}}), the time derivative of (\mathbbf{\ref{lyapunov}}) yields
\begin{align}
\dot{V} &\leq -\frac{1}{2}\xi^{T}\Upsilon\xi+ \Gamma+ s^{T} \lbrace-(c+\rho+\beta)\frac{s}{| s|}+\sigma \rbrace +||s||\Gamma_1 \nonumber\\
& \qquad \qquad - \sum_{i=0}^{1}(\hat{\gamma}_i-\gamma^{*}_i)||\xi||^i |s| + \frac{1}{\alpha_2}\hat{\gamma}_2 \dot{\hat{\gamma}}_2+ \Gamma \beta \dot{\beta}+\Gamma_1 \rho \dot{\rho} \nonumber\\
& \leq -\frac{1}{2} \lambda_{\min}(\Upsilon)||\xi||^2  +f(||\xi||)\label{case 2} 
\end{align}
where $$f(||\xi||)= -\varsigma \underline{\gamma}_2||\xi||^3 +\lbrace(1+\delta \rho_0)\Gamma_1+2\gamma^{*}_0\rbrace ||B^TP||||\xi|| + 2\gamma^{*}_1||B^TP||||\xi||^2 + (1+ \delta \beta_0) \Gamma $$ denotes a $3^{rd}$ degree polynomial of $||\xi||$. Now, applying Descartes' rule of sign change, it can be noticed that $f(||\xi||)=0$ has exactly one positive real root. So, other two roots are either two negative real roots or a pair of complex conjugate roots. Let, $\iota > 0$ be the positive real root of $f(||\xi||)=0$. Moreover, since coefficient of the cubic term (the highest degree term) of $f(||\xi||)$ is negative, $f(||\xi||)<0$ $\forall ||\xi||> \iota$ (Bolzano's Intermediate Value Theorem). 

Following the lines of proof as in \cite{roy2017adaptive4}, UUB stability can be claimed when $||\xi||> \iota$. Further, since $\varsigma$ is a design parameter (defined in (\mathbbf{\ref{alpha1}})), one can select high value of $\varsigma$ which helps to reduce $\iota$ and achieve better tracking accuracy. 

However, the control law (\mathbbf{\ref{delta u2}}) is discontinuous in nature. So, it is modified as below with the help of a user-defined scalar $\epsilon \in \mathbb{R}^{+}$:
\begin{align}
&\Delta {u}(t) =- \begin{cases}
    \zeta\frac{s}{| s|}       &  \text{if } | s| \geq \epsilon\\
    \zeta\frac{s}{\epsilon}        &  \text{if } | s| < \epsilon\\
  \end{cases}\label{delta u21}
\end{align}
The modified closed-loop stability under (\mathbbf{\ref{delta u21}}) is stated below:
\begin{lemma}
The system (\mathbbf{\ref{sys}}) employing ARTDC with the switching control input (\mathbbf{\ref{delta u21}}) follows similar stability notion mentioned in Theorem 3.1 when $|s| \geq \epsilon$, while the system remains UUB when $|s| < \epsilon$.
\end{lemma}
\textit{Proof.}
For the condition $|s| < \epsilon$, the following two possible conditions are identified:
\par \textbf{Case (iii):} $ | s | < \epsilon \land \lbrace \hat{\gamma}_i \leq \underline{\gamma}_i \lor  s \dot{s} >0,   i=0,1,2 \rbrace $ 
\par \textbf{Case (iv):} $| s | < \epsilon \land \lbrace s \dot{s} \leq 0 \lor \beta \leq \underline{\beta} \lor \rho \leq \underline{\rho} \rbrace $
\par The stability aspect of these two cases are analysed using (\mathbbf{\ref{delta u2}})-(\mathbbf{\ref{beta1}}) and, the results obtained from Lemma 3.3 and Theorem 3.1.
\par \textbf{Case (iii):} $ | s | < \epsilon \land \lbrace {\alpha} \leq \gamma \lor  s \dot{s} >0, i=0,1,2 \rbrace$ \\
Following the lines of proof as in \cite{roy2019overcoming}, it can be shown that $\exists {\gamma}_{iM}$ such that $\hat{\gamma}_i \leq {\gamma}_{iM}$ $\forall i=0,1,2$ for this case. Using (\mathbbf{\ref{delta u2}}) and the fact $| s | < \epsilon$, for this case we have
\begin{align}
\dot{V} &\leq -\frac{1}{2}\xi^{T}\Upsilon\xi+ \Gamma+ s\lbrace-(c+\rho+\beta)\frac{s}{\epsilon}+\sigma \rbrace+|s|\Gamma_1 \nonumber\\
& \qquad \qquad + \sum_{i=0}^{1}(\hat{\gamma}_i-\gamma^{*}_i)||\xi||^i |s| + \frac{1}{\alpha_2}\hat{\gamma}_2 \dot{\hat{\gamma}}_2+ \Gamma \beta \dot{\beta}+\Gamma_1 \rho \dot{\rho} \nonumber\\
& \leq -\frac{1}{2} \lambda_{\min}(\Upsilon)||\xi||^2+\epsilon \gamma_{1M} ||\xi||+ \epsilon (\gamma_{0M}+\gamma_{2M}). \label{case 3}
\end{align}

\par \textbf{Case (iv):} \quad $| s | < \epsilon \land \lbrace s \dot{s} \leq 0 \lor \beta \leq \underline{\beta} \lor \rho \leq \underline{\rho} \rbrace$\\
Similarly, for this case
\begin{align}
\dot{V} &\leq -\frac{1}{2}\xi^{T}\Upsilon\xi+ \Gamma+ s\lbrace-(c+\rho+\beta)\frac{s}{\epsilon}+\sigma \rbrace+||s||\Gamma_1 \nonumber\\
& \qquad \qquad - \sum_{i=0}^{1}(\hat{\gamma}_i-\gamma^{*}_i)||\xi||^i |s| + \frac{1}{\alpha_2}\hat{\gamma}_2 \dot{\hat{\gamma}}_2+ \Gamma \beta \dot{\beta}+\Gamma_1 \rho \dot{\rho} \nonumber\\
& \leq-\frac{1}{2} \lambda_{\min}(\Upsilon)||\xi||^2  -\varsigma \underline{\gamma}_2 ||\xi||^3 +\lbrace(1+\delta \rho_0)\Gamma_1+2\gamma^{*}_1\rbrace \epsilon||\xi||+ (1+ \delta \beta_0)\Gamma +2\epsilon \gamma^{*}_0   \label{case 4}
\end{align}
Case (iv) can be analysed exactly like Case (ii). With similar arguments as in Sect. II.A of \cite{roy2019overcoming}, UUB can be claimed for closed-loop system when $|s| < \epsilon$. 

\begin{remark}[Independence from the variation of input delay]
The closed-loop stability analysis reveals that, utilizing ARTDC, no restriction is imposed on the rate of change of delay, i.e.,  $\dot{h}_{in}$ and $\ddot{h}_{in}$. This allows ARTDC to tackle arbitrary variation of input delay. 
\end{remark}
\begin{remark}[Importance of $\rho$ and $\beta$]\label{rho_beta}
It can be noted from (\mathbbf{\ref{case 1}}), that the negative semi-definite result was achieved by cancelling the terms $\Gamma$ and $||s|| \Gamma_1$ via $\beta$ and $\rho$ respectively. Compared to \cite{Ref:self}, such design helps ARTDC to provide asymptotic convergence of tracking error compared to the UUB result obtained in the former, when tracking error moves away from switching surface (i.e., in Case (i)). 
\end{remark}
\begin{remark}
It is to be noted that, the various terms like $\gamma^{*}_0,\gamma^{*}_1, \Gamma, \Gamma_1$ and ${\gamma}_{iM}, i=0,1,2$ are only used for analytical purpose and they are not utilized for controller design. Further, high values of $\alpha_i, i=0,1,2 $ helps to achieve better error convergence. On the other hand, (\mathbbf{\ref{case 2}}) and (\mathbbf{\ref{case 4}}) reveal that smaller values $\delta, \beta_0$ and $ \rho_0$ reduce the error bounds in the UUB conditions. 
\end{remark}
\textit{\textbf{Comparison with the existing works:}}
In comparison with the existing works in literature, the advantages of the proposed ARTDC are summarized below:
\begin{itemize}
\item In contrast to the method AROLC as in \cite{Ref:self}, the proposed ARTDC framework extends the maximum allowable input delay of the system for same choice of \textit{nominal control parameters} according to Lemma 4. 
\item Unlike \cite{Ref:31, Ref:32, Ref:33}, ARTDC does not depend on $\dot{h}_{in}$ and $\ddot{h}_{in}$ which allows the system to be insensitive towards arbitrary variation of input delay. Specifically, fast variation of input delay is allowed by ARTDC compared to only slow variation allowed in those works. Moreover, ARTDC does not involve any threshold value in its adaptive laws (\mathbbf{\ref{alpha0}})-(\mathbbf{\ref{beta1}}) and allows the switching gains to decrease when error decreases (i.e. when $s \dot{s} \leq 0 $). Hence, ARTDC avoids over- and under-estimation problem of switching gain.
\item ARTDC is specifically designed by considering a more generalized structure of the upper bound of uncertainty (as provided in (\mathbbf{\ref{n sigma}})) which involves system states. Hence, ARTDC removes the conservativeness regarding presumption of constant upper bound on uncertainties like \cite{Ref:self}. 
\item While \cite{Ref:self} provided UUB results for all the conditions, an asymptotic convergence of tracking error $\xi$ is achieved for Case (i) by the proposed ARTDC. This is due the introduction of the gains $\rho$ and $\beta$. This case is important as it signifies the situation when error is increasing and going away from the switching surface $s$. Hence, improved tracking accuracy is obtained in ARTDC. 
Moreover, with the modified adaptive law of ARTDC, the Case (iii) of \cite{Ref:self}, which shows UUB result, is subsumed by ARTDC in Case (i). This, further, allows ARTDC to provide better tracking accuracy through asymptotic convergence of error. 
\end{itemize}

\section{Simulation Results}
In this section, the effectiveness of the proposed ARTDC scheme is verified for an SBW system with the same parameters as in the previous chapter.

\begin{figure}[h!]
      \centering
     \includegraphics[width=3.6 in,height=1.5 in]{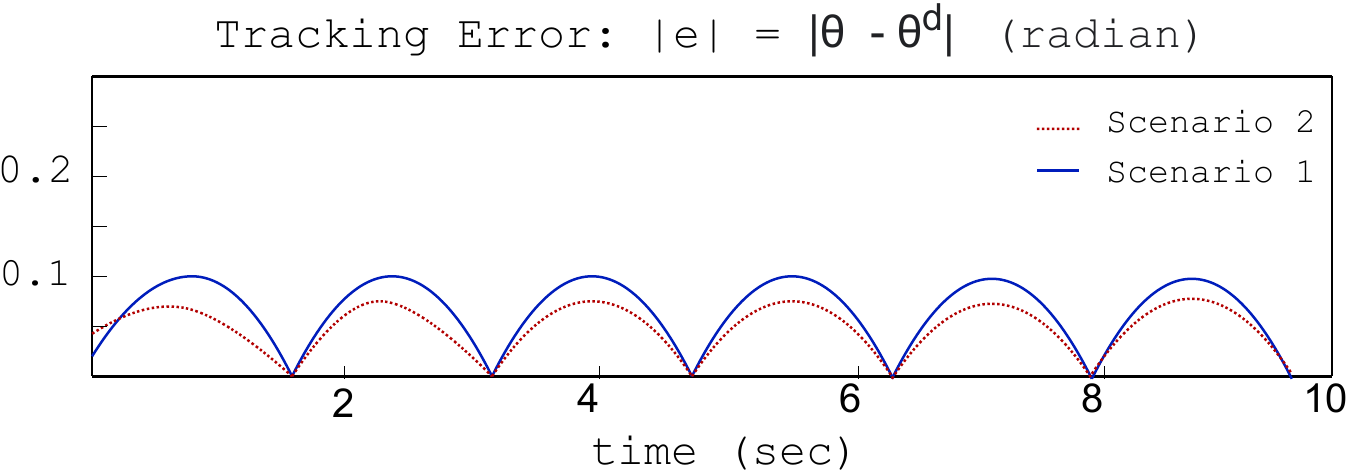}
      \caption{Trajectory tracking performance comparison of ARTDC under various scenarios.}
      \label{fig:1_1}
\end{figure}

\begin{figure}[h!]
      \centering
      \includegraphics[width=3.3 in,height= 2.3 in]{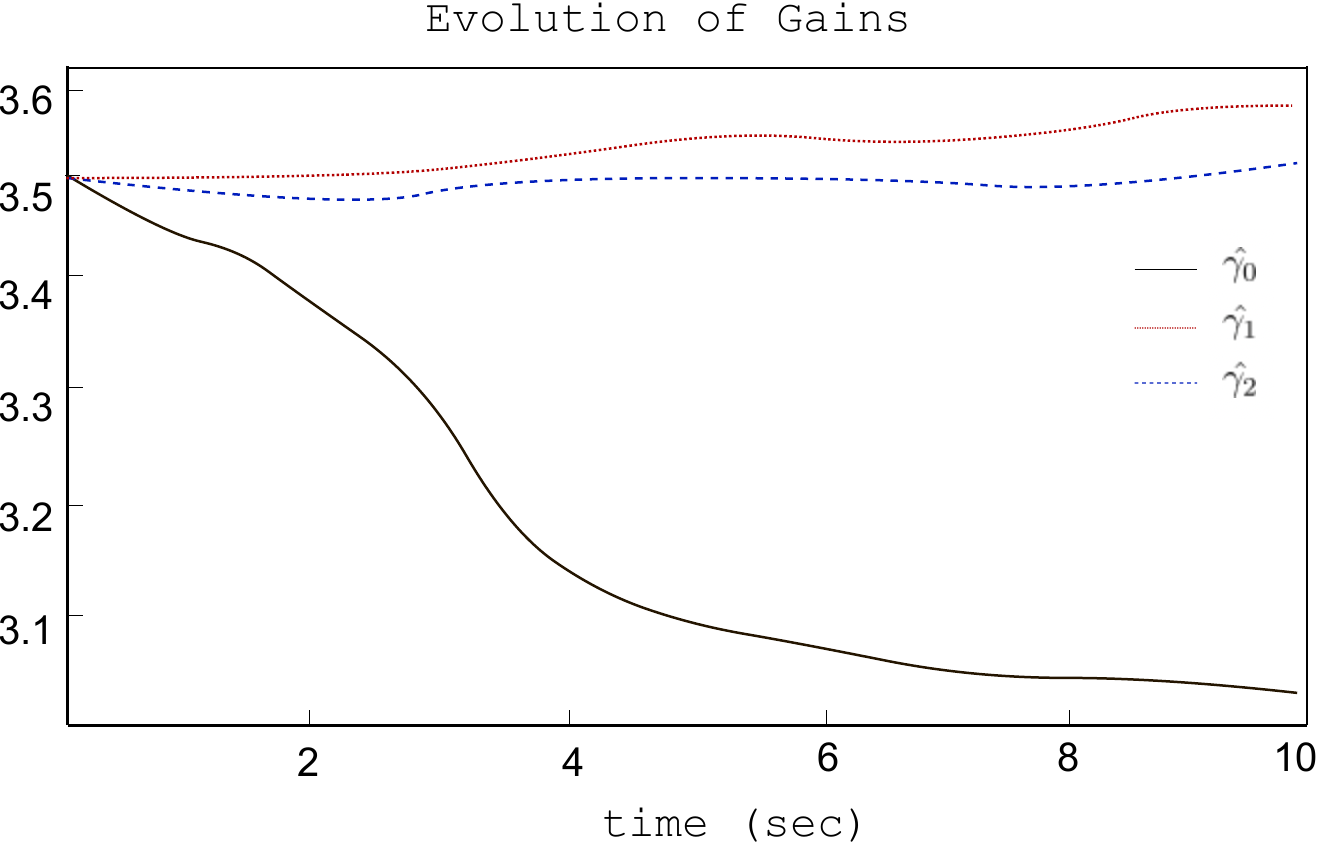}
      \caption{Evolution of adaptive gains.}
      \label{fig:3_1}
\end{figure}

\begin{figure}[h!]
      \centering
      \includegraphics[width=3.5 in,height=2.3 in]{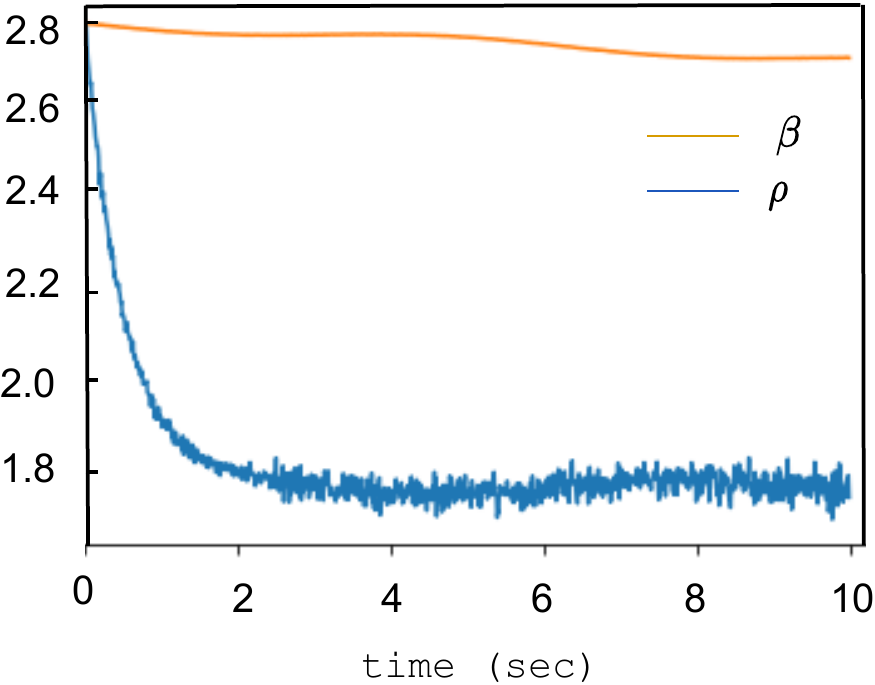}
      \caption{Evolution of Gains $\beta$ and $\rho$}
      \label{fig:4_1}
\end{figure}

\begin{figure}[h!]
      \centering
       \includegraphics[width=4 in,height=2.6 in]{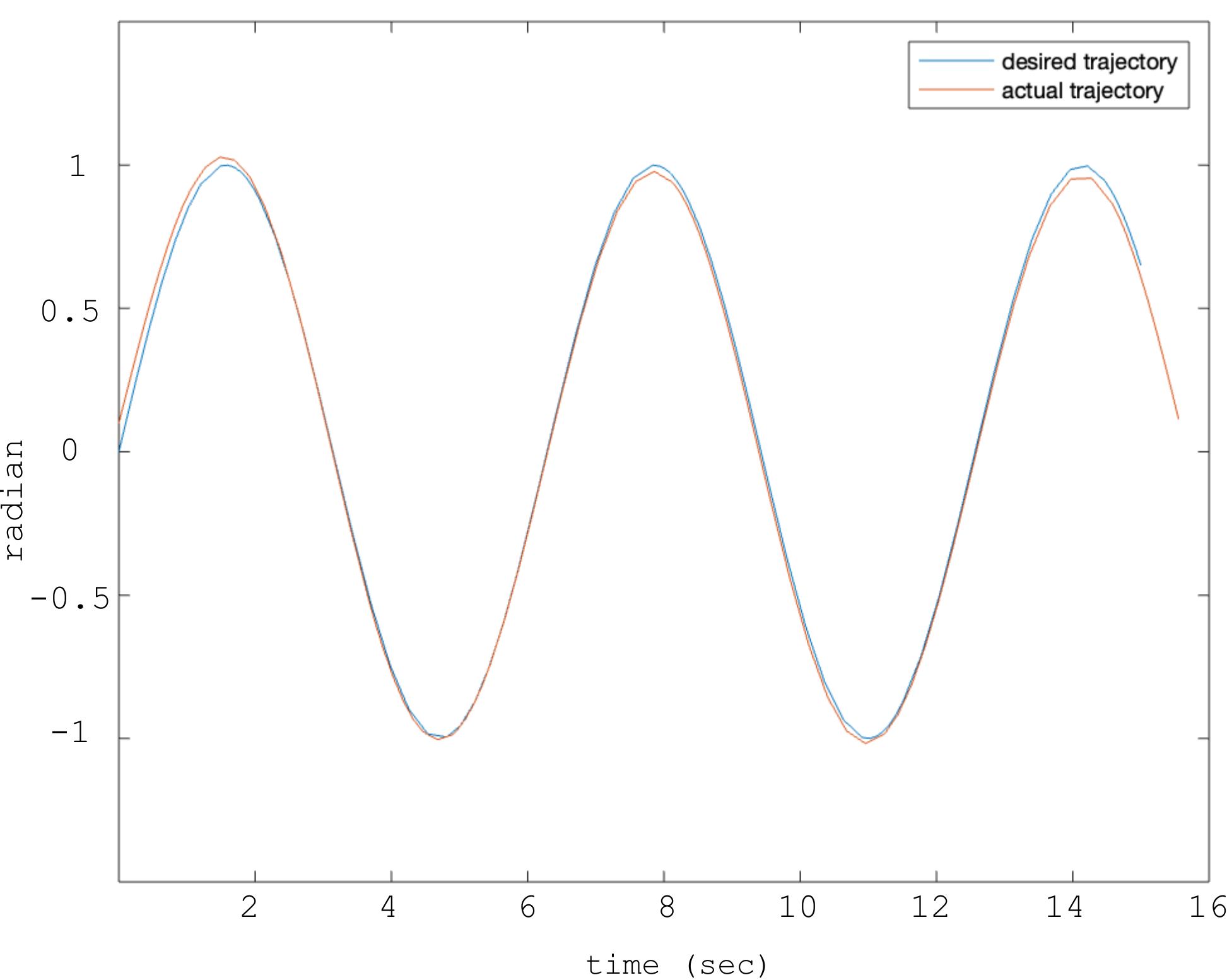}
      \caption{State response with ARTDC}
      \label{fig:2_1}
\end{figure}

 We apply the proposed ARTDC laws (\mathbbf{\ref{input1}}), (\mathbbf{\ref{aux2}}) and (\mathbbf{\ref{delta u2}}) with adaptive laws (\mathbbf{\ref{alpha0}})-(\mathbbf{\ref{beta1}}) using the following control design parameters: $Q=I, K=1,\Omega=0.5, r=1.01, \eta=0.7$, $\bar{g}=0.5$, $\delta=10, \underline{\rho}=\underline{\beta}=0.05, \varsigma=0.1, {\alpha}_0={\alpha}_1=0.82, {\alpha}_2=1, \underline{\gamma}_i=0.001$, $\hat{\gamma}_{i0}=3$ $\forall i=0,1,2$ and $\epsilon=0.1$ and $\beta = \rho = 2.8$. Such selections of parameters satisfy the conditions (\mathbbf{\ref{mass}}) and (\mathbbf{\ref{delay value}}). The objective is to track a desired trajectory $q^d=\sin(t)$ under the assumption that the input has time-delay and is governed by the equation, $h(t)= 0.02|sin(0.01t)|$. 
 
To verify the effectiveness of various adaptive-robust gains we select the following two scenarios:
\begin{itemize}
\item Scenario 1: $\hat{\gamma}_1=\hat{\gamma}_2=\beta=\rho=0$ in (\ref{delta u2}); 
\item Scenario 2: $\hat{\gamma}_j$, $j=0,1,2$ and $\beta,\rho$ are evaluated as in (\ref{delta u2}) and (\ref{alpha0})-(\ref{beta1}).
\end{itemize}
Scenario 1 can be considered to be equivalent to that of AROLC \cite{Ref:self}.
Figure \ref{fig:1_1} demonstrates the benefit of additional adaptive-robust gains as in (\ref{c}), stemming from the state-dependent uncertainty structure as in Lemma 1, where ARTDC in Scenario 2 clearly provides significant performance improvement. Further, evolution of various adaptive-robust gains for the proposed design are depicted in Figs. \ref{fig:3_1} and \ref{fig:4_1}. The state response for the proposed controller against the desired trajectory is shown in Fig. \mathbbf{\ref{fig:2_1}}. These plots demonstrate the benefits of additional adaptive-robust gains as in (\mathbbf{\ref{c}}), stemming from the state-dependent uncertainty structure considered by the proposed method. 

\chapter{Conclusions and Future Work}
\label{ch:conc}
In this thesis, we designed and presented controllers for tackling two different aspects of steer-by-wire systems. A new concept of adaptive control design for steer-by-wire systems was introduced to tackle state-dependent uncertainties which are not a priori bounded. The proposed design did not require any knowledge of structure and bound of uncertainties. The closed-loop stability was analytically established to be uniformly ultimately bounded. The effectiveness of the proposed concept was validated via simulations in comparison with the state of the art. 

In the second part, a new adaptive-robust tracking controller was proposed to negotiate the influences of time varying input delay and state-dependent uncertainty simultaneously in SBW systems. In contrast to the existing adaptive-robust strategies, (i) the proposed framework avoided any restrictive assumption such as uncertainties being upper bounded by a constant a priori: instead, the control structure relied on a linear-in-parameters upper bound structure of uncertainty, stemming from physical properties of a system; (ii) in addition, increased delay tolerability and (partially) improved stability result were also achieved. 

While tackling the input delay, the proposed work still requires nominal knowledge of system parameters. Therefore, an important future work would be to eliminate/minimize such requirements. Further, we would attempt to verify the proposed algorithms on real-life vehicles in future.


\chapter*{Related Publications}
\label{ch:relatedPubs}
The work presented in this thesis resulted in the following two publications:
\begin{itemize}
    \item \textbf{H. Shukla}, S. Roy and S. Gupta, 'Robust Adaptive Control of Steer-by-wire Systems Under Unknown State-dependent Uncertainties,'
    International Journal of Adaptive Control and Signal Processing, 2021. 
    \item S. Roy and \textbf{H. Shukla}, "Adaptive-Robust Controller for a Class of Systems with Time Varying Input Delay and State-Dependent Uncertainty," In The Book: `Control Strategy for Time Delay Systems-Part I: Theory and Concepts', Elsevier, pp. 241-271, 2020.
\end{itemize}

\bibliographystyle{IEEEtran}
\bibliography{sampleBib} 

\end{document}